\documentclass{article}
    \PassOptionsToPackage{numbers, compress}{natbib}

\usepackage[main, final]{neurips_2026}

\usepackage{amsmath,amsfonts,bm}

\def\eqref#1{equation~\ref{#1}}

\def\1{\bm{1}}

\DeclareMathAlphabet{\mathsfit}{\encodingdefault}{\sfdefault}{m}{sl}
\SetMathAlphabet{\mathsfit}{bold}{\encodingdefault}{\sfdefault}{bx}{n}

\usepackage{colortbl}
\usepackage{color}
\usepackage[dvipsnames]{xcolor} 
\usepackage[colorlinks=True,
            linkcolor=BrickRed,      
            anchorcolor=BrickRed, 
            citecolor=OliveGreen,   
            ]{hyperref}
\usepackage{graphicx} 
\usepackage{wrapfig}
\usepackage{amsmath,amssymb,amsthm}

\usepackage[utf8]{inputenc} 
\usepackage[T1]{fontenc}    
\usepackage{url}            
\usepackage{booktabs}       
\usepackage{amsfonts}       
\usepackage{nicefrac}       
\usepackage{microtype}      
\usepackage{xspace}
\usepackage{url}
\usepackage{fontawesome5}
\usepackage{mdframed}
\usepackage{tikz}
\usepackage{multirow}
\RequirePackage[export]{adjustbox}
\usepackage{pifont}
\usepackage{cleveref}
\usepackage{float}
\usepackage{placeins}
\usepackage{mathtools}
\usepackage{enumitem}
\usepackage{subcaption}

\theoremstyle{plain}
\newtheorem{theorem}{Theorem}
\newtheorem{lemma}{Lemma}

\newtheorem{corollary}{Corollary}

\theoremstyle{definition}
\newtheorem{definition}{Definition}
\newtheorem{assumption}{Assumption}

\theoremstyle{remark}
\newtheorem{remark}{Remark}

\usepackage[ruled,vlined]{algorithm2e} 
\usepackage{amsmath, amssymb, bm} 
\usetikzlibrary{fit,calc}
\definecolor{yellow}{rgb}{1, 0.8, 0.2}
\definecolor{green}{rgb}{0.5, 0.8, 0.5}
\definecolor{blue}{rgb}{0.4, 0.7, 1}

\DeclarePairedDelimiterX{\infdivx}[2]{(}{)}{%
  #1\;\delimsize\|\;#2%
}

\newcommand{\themodel}{\textbf{\texttt{BindMol}}\xspace}
\crefname{algocf}{Algorithm}{Algorithms}

\title{Binding Mode Matters: Hotspot-Aware Drug Discovery via Explorative Preferences}

\author{%
\normalsize
Dingshuo Chen\textsuperscript{\textmd{1,2,4}}\quad
Hao Yang\textsuperscript{\textmd{3}}\quad
Kuangqi Zhou\textsuperscript{\textmd{4}}\quad
Zhixun Li\textsuperscript{\textmd{5}}\quad\\
\normalsize
\textbf{Qiang Liu}\textsuperscript{\textmd{1,2}}\quad
\textbf{Xiangxiang Zeng}\textsuperscript{\textmd{6}}\quad\
\textbf{Shu Wu}\textsuperscript{\textmd{1,2}\thanks{Corresponding author: Yaosen Min (\texttt{yaosenmin@zgci.ac.cn}), Shu Wu (\texttt{shu.wu@nlpr.ia.ac.cn}).}}\quad
\textbf{Yaosen Min}\textsuperscript{\textmd{4}\footnotemark[1]}\\
\textsuperscript{\textmd{1}} Institute of Automation, Chinese Academy of Sciences\\
\textsuperscript{\textmd{2}} School of Artificial Intelligence, University of Chinese Academy of Sciences \\
\textsuperscript{\textmd{3}}Shanghai Institute for Advanced Immunochemical Studies, ShanghaiTech University\\
\textsuperscript{\textmd{4}} Zhongguancun Academy \& Zhongguancun Institute of Artificial Intelligence\\
\textsuperscript{\textmd{5}} The Chinese University of Hong Kong\\
\textsuperscript{\textmd{6}}College of Computer Science and Electronic Engineering, Hunan University\vspace{-1em}
}

\begin{document}

\maketitle

\begin{abstract}
  Discovering hit molecules requires not just high binding affinity, but also the identification of diverse binding modes, which are critical for experimental assays. Existing generative approaches have predominantly relied on optimizing a scalar docking score, obscuring the distinct contributions of key binding determinants.
  To this end, we introduce a paradigm shift by formulating target-based drug design as a multi-objective exploration task, where each objective explicitly corresponds to enhancing interactions with a specific hotspot. 
  Here, we introduce \textbf{\themodel}, a novel generative framework driven by a customized multi-objective reinforcement learning algorithm. By incorporating explorative preferences during training, our approach efficiently uncovers molecules with diverse and desirable binding profiles.
  Empirical results demonstrate that \textbf{\themodel} facilitates the discovery of high-affinity compounds characterized by both structural novelty and diverse binding modes. Validated across target-based drug discovery and multi-property optimization tasks, our approach provides a versatile paradigm for goal-oriented drug discovery.
\end{abstract}

\section{Introduction}

Target-based drug design aims to identify ligands that not only bind to a target protein with high affinity but also engage specific residues to trigger a desired pharmacological response \citep{ferreira2015molecular}. The recent surge in deep generative models has accelerated this process, offering a powerful toolkit that spans \emph{de novo discovery} and \emph{goal-oriented generation} \cite{du2024machine}. Reinforcement learning, in particular, has emerged as a dominant paradigm for conditional generation \citep{olivecrona2017molecular, lee2024molecule} and optimization \citep{zhou2019optimization, yang21freed}. It generates molecular structures by maximizing a reward reflecting drug-likeness and bioactivity \citep{olivecrona2017molecular}.

\begin{wrapfigure}{r}{.5\textwidth}
\vspace{-2mm}
\centerline{
\begin{tabular}{c}
\hspace*{-3mm}\includegraphics[width=.5\textwidth,height=!]{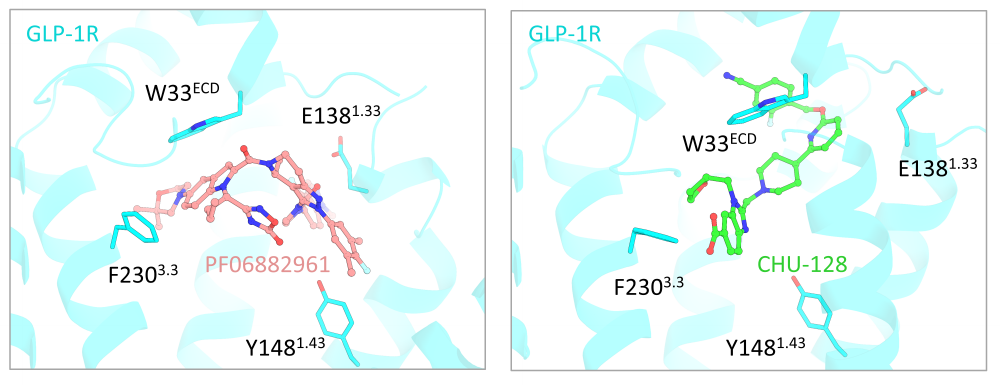}
\end{tabular}}
\vspace{-3mm}
\caption{Illustration of two agonists of GLP-1R, highlighting their distinct binding modes within the same protein pocket.}
\vspace{-3mm}
\label{fig:glp1}
\end{wrapfigure}
Despite these advances, a misalignment persists between the complexity of biochemical interactions and the simplified objectives used in RL training. Most existing approaches reduce the ligand-protein interaction to a single scalar metric, such as the docking score. While computationally convenient, such scalar measures often overlook the complexity of interactions: \textbf{\emph{compounds targeting the same protein can adopt different binding modes, leading to distinct functional and selectivity profiles}}. For instance, two well-known GLP-1 agonists exhibit similar affinities but engage distinct sub-regions, resulting in either broad-spectrum or biased signaling \citep{zhang2020differential}, as illustrated in \Cref{fig:glp1}. 


To address this limitation, we reformulate target-based drug discovery as a multi-objective optimization problem, where each objective corresponds to a distinct \emph{hotspot} within the binding pocket. This conceptual shift naturally encourages the exploration of diverse binding modes by explicitly exposing trade-offs between interactions with different pocket regions.
It also provides a more robust alternative to scalar objectives, which are known to be prone to reward hacking and suboptimal local solutions \citep{fromer2023computer}.
Specifically, we propose a novel reinforcement learning framework for drug discovery, termed \textbf{\themodel}.
Its realization relies on three parts that bridge chemical validity, biological specificity, and algorithmic efficiency.

We explore the discrete chemical space using a fragment-based action space \citep{yang21freed}, where molecules are incrementally assembled in a \emph{Lego-like} fashion. A compact yet informative fragment vocabulary is obtained via an energy-based model trained with noise contrastive estimation \citep{gutmann2010noise}.
We theoretically ground fragment selection as a stability analysis problem on a thermodynamic potential surface, where high-scoring fragments act as stabilizing anchors for the ligand-protein complex.
With a robust action space established, we next focus on the reward signal. Instead of flattening complex biological interactions into a docking score, we introduce a hotspot-aware vectorized reward by clustering the binding pocket into distinct \emph{hotspots} based on reference structures, with each hotspot defining a separate objective. 
This design compels the agent to navigate trade-offs across sub-pockets, naturally leading to the discovery of diverse binding modes.

However, maximizing a vectorized reward is far more challenging than a scalar one \citep{lu2023multi, alegre2023sample}. To enable precise control over multi-objective trade-offs, we establish \emph{Envelope SAC}, a principled actor-critic framework that unifies the optimization of preference-aware critics and policies. This formulation significantly advances the theoretical scope of the envelope operator \citep{yang2019generalized}, moving beyond its value-based origins to the continuous domain. By constructing a utility envelope that aligns vectorized Bellman updates, our method efficiently approximates the convex hull of the Pareto frontier, ensuring robust performance across diverse preferences.

We empirically validate \themodel on binding affinity optimization tasks across five protein targets and the extensive GuacaMol benchmark. Our results demonstrate that \themodel not only generates valid, high-affinity compounds but also successfully discovers diverse binding modes that cover a broader range of the bioactive chemical space compared to state-of-the-art baselines. The main contributions of this work are three-fold:

\ding{182} We propose the first molecular generation framework that treats binding-mode diversity as a \emph{first-class} modeling objective, explicitly accounting for biologically grounded structural variations. This moves target-based design one step closer to treating molecular generation as an exploration of protein--ligand interaction patterns.

\ding{183} We propose \themodel, a principled multi-objective reinforcement learning framework that integrates energy-based fragment scoring with hotspot-aware vectorized rewards, and develops Envelope SAC, an actor--critic algorithm capable of efficiently approximating the Pareto frontier under preference-conditioned optimization.

\ding{184} We demonstrate the superiority of \themodel across diverse benchmarks. It outperforms existing methods in generating high-affinity ligands with novel binding modes across five protein targets and achieves state-of-the-art results on the comprehensive GuacaMol multi-objective benchmark.

\section{Related Work}

\paragraph{Fragment-based molecular generation}
Fragment-based molecular generation has emerged as an effective paradigm for drug discovery. Methods such as MARS \citep{xie2021mars}, FREED \citep{yang21freed}, and MoLeR \citep{maziarz2021learning} assemble molecules from fragments to efficiently explore chemical space. Related approaches exploit informative substructures, including principal subgraph mining \citep{kong2022molecule} and molecular rationales \citep{jin2020multi}, to guide multi-objective generation. Incorporating protein structural context further improves generation quality, for example through transferable subpocket prototypes \citep{zhang2023learning} or pocket-aware fragment assembly via diffusion \citep{ghorbani2023autoregressive}. Beyond generation, HN-GFN \citep{zhu2023sample} enables sample-efficient multi-objective optimization, and recent antibiotic discovery results \citep{krishnan2025generative} demonstrate the practical impact of fragment-based strategies.

\paragraph{RL-driven molecular generation}
Deep reinforcement learning has been extensively applied to molecular design using both sequence- and graph-based formulations. Sequence-based policy optimization methods, including REINVENT \citep{olivecrona2017molecular} and its variants with curriculum learning \citep{guo2022improving}, adapt SMILES generators toward target properties. Graph-based approaches \citep{zhou2019optimization} instead operate directly on molecular structures through chemistry-aware edit actions, often without pretraining. Fragment-based RL methods further address exploration efficiency and chemical validity by assembling molecules from learned vocabularies \citep{yang21freed, lee2024molecule}. More recently, structure-aware techniques such as AliDiff \citep{gu2024aligning} explicitly align generative models with protein-binding preferences. Together, these works demonstrate the flexibility of RL for molecular design, while highlighting persistent challenges related to reward design and exploration.

\paragraph{Multi-objective reinforcement learning}
Multi-objective reinforcement learning (MORL) spans both tabular and deep settings, with single-policy and multi-policy formulations \citep{felten2023toolkit}. Tabular methods either optimize scalarized objectives \citep{van2013scalarized} or maintain non-dominated value estimates, as in Pareto Q-learning \citep{van2014multi}. In the deep setting, single-policy approaches such as EUPG \citep{roijers2018multi} optimize user-defined utilities, while multi-policy methods dominate, including weight-conditioned models \citep{yang2019generalized, lu2023multi}, evolutionary approaches \citep{xu2020prediction}, return-conditioned models \citep{reymond2022pareto}, and GPI frameworks \citep{alegre2023sample}. These methods differ mainly in how preferences are incorporated and how the Pareto front is approximated.

\begin{figure*}[t]
    \centering
    \includegraphics[width=\textwidth,height=!]{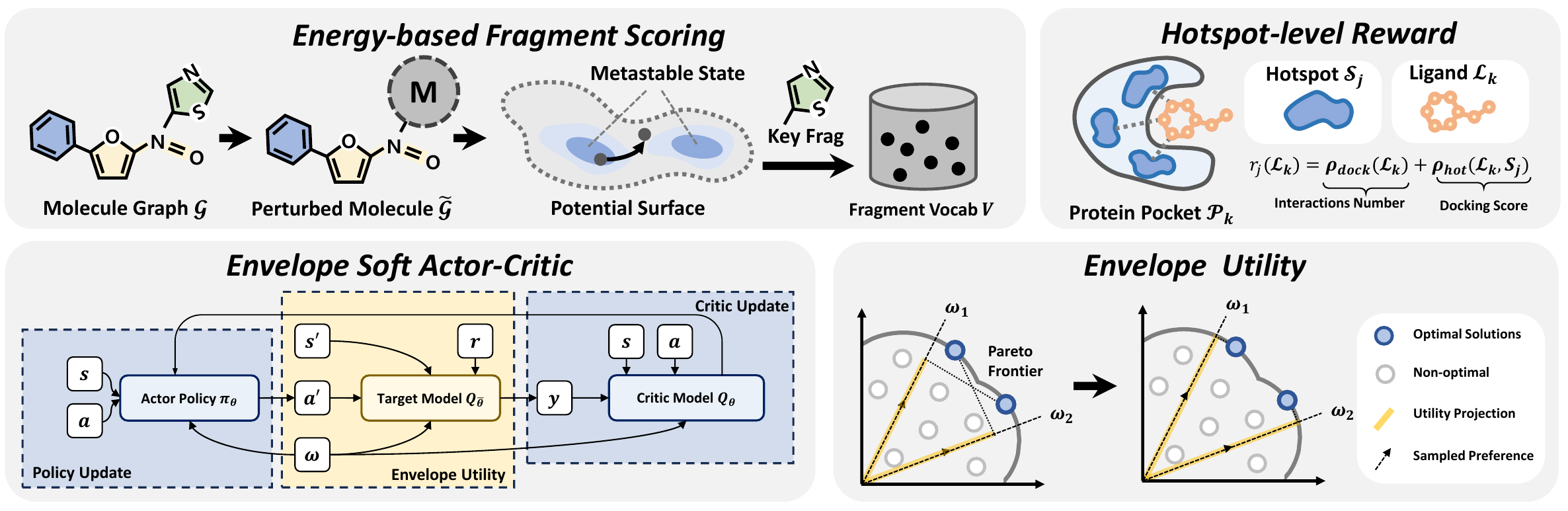}
    \vspace{-5mm}
    \caption{Overview of the \themodel consisting of three components: energy-based fragment scoring, Envelope SAC, and hotspot-level reward design. \textbf{(Top left)} Energy-based fragment scoring selects stabilizing fragments by an energy-based model trained via noise contrastive estimation. \textbf{(Bottom left)} Overview of the Envelope SAC workflow (double $Q_{\theta}$ models omitted for clarity). \textbf{(Top right)} Hotspot-aware rewards are computed from the sum of interaction counts and docking scores, followed by normalization. \textbf{(Bottom right)} Illustration of envelope utility: unlike scalarized RL without maximization over preference, our method identifies the optimal utility for each preference.}
    \label{fig:framework}
    \vspace{-1mm}
\end{figure*}

\section{Preliminaries}
\paragraph{Markov decision process.}
We formalize a multi-objective reinforcement learning problem as a Multi-Objective Markov Decision Process (MOMDP), defined by
\[
\mathcal{M} = \langle \mathcal{S}, \mathcal{A}, \mathcal{P}, \mathbf{r}, \Omega \rangle
\]
with state space $\mathcal{S}$, action space $\mathcal{A}$, transition probability distribution $\mathcal{P}(s'|s,a)$, vector reward function $\mathbf{r}$, and the space of preference $\Omega$.
If $\bm \omega \in \Omega$ is fixed to a single value, this MOMDP collapses into a standard MDP. On the other hand, if we consider all possible returns from an MOMDP, we have a Pareto frontier $\mathcal F^* := \{\bm {\hat r} \mid \not\exists \bm {\hat r}'\geq \bm {\hat r}\}$, 
where the return $\bm {\hat r} := \sum_t\gamma^t \bm r(s_t,a_t)$.

\paragraph{Fragment-based molecular generation}
\label{sec:prelim_generator}

Following \citet{yang21freed}, each transition is driven by a triplet action $\mathbf{a}_t = (u_t, f_t, v_t)$, which attaches a chosen fragment $f_t$ from the vocabulary to the current graph $\mathcal{G}_t$ at site $u_t$, using the fragment's site $v_t$. The generation policy $\pi_\theta(\mathbf{a}_t | s_t)$ is factorized into three hierarchical sub-policies:
\begin{equation}
    \pi_\theta(\mathbf{a}_t \mid \mathcal{G}_t) = 
    \mathbb{P}(u_t \mid \mathcal{G}_t) \cdot
    \mathbb{P}(f_t \mid u_t, \mathcal{G}_t) \cdot
    \mathbb{P}(v_t \mid f_t, u_t, \mathcal{G}_t).
    \label{eq:policy_factorization}
\end{equation}
Detailed descriptions of the molecular generator are provided in Appendix~\ref{app:generator}.

\section{Method}
The three components of \themodel address complementary stages of one design problem. Energy-based fragment scoring (\Cref{subsec:ebm}) constructs a compact, target-adapted molecular search space; hotspot-aware vector rewards (\Cref{subsec:reward}) formulate preference-controllable exploration of binding modes; and Envelope SAC (\Cref{subsec:env_sac}) optimizes the resulting conflicting objectives in the combinatorial generation process. The overall framework is illustrated in \Cref{fig:framework}.

\subsection{Energy-Based Fragment Scoring}
\label{subsec:ebm}

We formulate the identification of goal-aware fragments as a stability analysis problem within a learned thermodynamic potential. We hypothesize that high-activity molecules represent metastable states in a joint structure-activity landscape \citep{bussi2020using}. Consequently, critical substructures act as stabilizing anchors; their perturbation should yield a significant energy fluctuation, pushing the system towards an unstable manifold.

Let the joint state space be $\mathcal{Z} = \mathcal{G} \times \mathcal{Y}$, where $\mathcal{G} = (\mathcal{V}_g, \mathcal{E}_g)$ denotes the molecular graph with node feature matrix $\mathbf{H} \in \mathbb{R}^{|\mathcal{V}_g| \times d}$, and $y \in \mathcal{Y} \subset \mathbb{R}$ is the target property. We model the compatibility probability density as a Boltzmann distribution $p_\theta(\mathbf{z}) \propto e^{-E_\theta(\mathbf{z})}$, governed by a parameterized energy function $E_\theta: \mathcal{Z} \to \mathbb{R}$ \citep{lecun2006tutorial}.
The energy architecture employs a message-passing encoder $\Phi_G: \mathcal{G} \to \mathbb{R}^h$ and a scalar projection $\Phi_y: \mathcal{Y} \to \mathbb{R}^h$. Their shared output width is a compact implementation choice, not a mathematical requirement: $\Phi_G$ summarizes the graph, whereas $\Phi_y$ lifts the scalar property into a conditioning representation, and different widths could also be concatenated. The scalar potential is derived via a fused non-linear map $\Psi$:
\begin{equation}
    E_\theta(\mathcal{G}, y) \coloneqq \Psi \left( \Phi_G(\mathbf{H}, \mathcal{E}_g) \, \| \, \Phi_y(y) \right),
\end{equation}
where $\|$ denotes concatenation. To ensure the physical validity of the landscape, we train $E_\theta$ using Noise Contrastive Estimation (NCE) \citep{gutmann2010noise}. We define the training objective $\mathcal{J}(\theta)$ to maximize the discriminative margin between the empirical distribution $\mathbb{P}_{\text{data}}$ and a generated noise distribution $\mathbb{P}_{\text{noise}}$:
\begin{equation}
    \mathcal{J}(\theta) = \mathbb{E}_{(\mathcal{G}, y) \sim \mathbb{P}_{\text{data}},(\tilde{\mathcal{G}}, \tilde{y}) \sim \mathbb{P}_{\text{noise}}} \left[ \log \left( 1 + e^{E_\theta(\mathcal{G}, y) - E_\theta(\tilde{\mathcal{G}}, \tilde{y})} \right) \right].
\end{equation}
Here, $\mathbb{P}_{\text{noise}}$ is constructed by stochastically corrupting either the topological structure $\mathcal{E}_g$ or the property label $y$, forcing the model to assign global minima only to chemically valid and property-matched pairs.
The empirical distribution may be formed from experimentally annotated target ligands or, for an underexplored target, from a public library scored by virtual screening. Thus, more target-relevant data improves the learned landscape, but the intended prerequisite is a reliable target structure and pocket definition rather than a large experimental activity set; a dataset-size analysis is provided in \Cref{app:data_sensitivity}.

Upon convergence, we utilize $E_\theta$ to quantify the contribution of substructures. For a molecule decomposed into a fragment set $\{\mathcal{F}_k\}_{k=1}^K$, let $\mathcal{P}_{\mathcal{F}_k^\perp}: \mathbb{R}^{|\mathcal{V}_g| \times d} \to \mathbb{R}^{|\mathcal{V}_g| \times d}$ be an orthogonal projection operator that effectively filters out the signal from fragments $\mathcal{F}_k$ (i.e., masking). The structural importance $\mathcal{I}(\mathcal{F}_k)$ is defined as the variation in potential energy under this projection:
\begin{equation}
\label{eq:perturbation}
\begin{split}
    \mathcal{I}(\mathcal{F}_k) & \coloneqq \delta E_\theta(\mathcal{G}, y; \mathcal{F}_k) = E_\theta(\mathcal{P}_{\mathcal{F}_k^\perp}(\mathbf{H}), \mathcal{E}_g, y) - E_\theta(\mathbf{H}, \mathcal{E}_g, y).
\end{split}
\end{equation}
A high value of $\mathcal{I}(\mathcal{F}_k)$ implies that the fragment $\mathcal{F}_k$ is essential for maintaining the low-energy state associated with high bioactivity.

\paragraph{Theoretical Understanding.}
The following result connects the energy change induced by fragment masking to the path-integrated gradient of the learned energy function, providing a principled fragment-importance signal rather than a purely heuristic score.

\begin{theorem}[Manifold Gradient Equivalence]
\label{thm:gradient_flow}
Let $\gamma: [0, 1] \to \mathbb{R}^{|\mathcal{V}_g| \times d}$ be a linear path parameterizing the feature ablation from $\mathbf{H}$ to $\mathcal{P}_{\mathcal{F}_k^\perp}(\mathbf{H})$. Assuming $E_\theta$ is continuously differentiable, the discrete energy variation $\mathcal{I}(\mathcal{F}_k)$ is exactly the path integral of the directional derivatives along $\gamma$:
\begin{equation}
    \mathcal{I}(\mathcal{F}_k) = - \int_{0}^{1} \left\langle \nabla_{\mathbf{H}} E_\theta(\gamma(t), y), \, \mathbf{1}_{\mathcal{F}_k} \odot \mathbf{H} \right\rangle_F \, dt,
\end{equation}
where $\langle \cdot, \cdot \rangle_F$ denotes the Frobenius inner product, $\mathbf{1}_{\mathcal{F}_k}$ is the indicator mask for fragment $\mathcal{F}_k$, and $\odot$ is the Hadamard product.
\end{theorem}

In essence, Theorem~\ref{thm:gradient_flow} characterizes the learned energy-gap score as an integrated sensitivity measure along the masking path. Detailed proofs, error analysis, and conditions for ranking consistency are provided in Appendix~\ref{sec:proofs}; pseudocode is given in \Cref{alg:frag_score}.

\subsection{Hotspot-Based Multi-Objective Reward Design}
\label{subsec:reward}

To guide the agent in exploring diverse binding modes rather than converging to a single dominant interaction pattern, we formulate the molecular generation task as a multi-objective optimization problem. We decompose the binding pocket into distinct spatial subpockets based on structural information derived from reference ligand-protein complexes from Protein Data Bank (PDB) \citep{berman2000protein}.

Let $\mathcal{D} = \{(\mathcal{P}_k, \mathcal{L}_k)\}_{k=1}^K$ denote a dataset of $K$ reference complexes for the target protein, where each entry consists of a protein structure $\mathcal{P}_k$ and its corresponding ligand $\mathcal{L}_k$. We represent the geometry of each residue $v$ by its side-chain centroid $\mathbf{c}_v$, and the ligand $\mathcal{L}_k$ by its set of heavy atom coordinates $\mathcal{A}(\mathcal{L}_k)$. For each complex, the binding site residue set $\mathcal{R}_k$ is identified via a centroid-to-atom distance threshold $\delta$ (e.g., $4\text{\AA}$):
\begin{equation}
    \mathcal{R}_k = \left\{ v \in \mathcal{P}_k \mid \min_{\mathbf{a} \in \mathcal{A}(\mathcal{L}_k)} \| \mathbf{c}_v - \mathbf{a} \|_2 < \delta \right\}.
\end{equation}
To maximize the distinctiveness of the objectives and reduce the bias toward common binding residues, we construct a \emph{non-overlapping residue set} $\mathcal{V}$ by excluding the intersection of all binding sites from their union:
\begin{equation}
    \mathcal{V} = \left( \bigcup_{k=1}^K \mathcal{R}_k \right) \setminus \left( \bigcap_{k=1}^K \mathcal{R}_k \right).
\end{equation}
Excluding the intersection does not discard conserved interactions from optimization: the global docking term below is shared by every objective and maintains pressure toward overall affinity. Instead, the hotspot-specific terms use differentiating residues; otherwise, conserved contacts can correlate all reward dimensions and let the policy improve every objective through the same anchor. This set $\mathcal{V}$ serves as the basis for subpocket identification. We partition it into $m$ spatial subpockets $\{\mathcal{S}_1, \dots, \mathcal{S}_m\}$ using spectral clustering \citep{von2007tutorial} on the Gaussian kernel of pairwise centroid distances. A full-union ablation is reported in \Cref{tab:full_union}, and clustering details are provided in \Cref{app:cluster}.

The reward function evaluates the generated molecule $\mathcal{M}$ based on both global binding affinity and local subpocket interactions. We define the reward vector $\bm r \in \mathbb{R}^m$, where the $j$-th component $r_j$ corresponds to the $j$-th subpocket $\mathcal{S}_j$ and consists of two terms:
$r_j(\mathcal{M}) = \rho_{\text{dock}}(\mathcal{M}) + \rho_{\text{hot}}(\mathcal{M}, \mathcal{S}_j).$
Here, $\rho_{\text{dock}}$ represents the docking score normalized to a fixed range (e.g., $[0, 1]$). Docking is performed once for each completed molecule, rather than after every fragment action; PLIP \citep{salentin2015plip} reuses that pose to compute $\rho_{\text{hot}}$ without another docking call.

Let $\eta(v, \mathcal{M})$ be the count of non-covalent interactions between molecule $\mathcal{M}$ and residue $v$. We first compute the average interaction density within the subpocket, denoted as $\bar{\eta}_j = \frac{1}{|\mathcal{S}_j|} \sum_{v \in \mathcal{S}_j} \eta(v, \mathcal{M})$. This value is then clipped and normalized to encourage valid contacts without over-saturation:
$\rho_{\text{hot}}(\mathcal{M}, \mathcal{S}_j) = \frac{1}{10} \min\left( \bar{\eta}_j, 10 \right).$
This formulation ensures that the agent optimizes for molecules that are not only energetically favorable globally but also form dense, specific interaction patterns with the target hotspots.

\subsection{Envelope Soft Actor-Critic}
\label{subsec:env_sac}

The reward formulation in Section \ref{subsec:reward} yields a vector $\bm r \in \mathbb{R}^m$, representing interactions across spatially disjoint subpockets. Reducing this vector to a static scalar via fixed weights prevents the agent from dynamically exploring trade-offs between different hotspots. To address this, we propose \emph{Envelope SAC}, which learns a universal policy capable of approximating the entire Pareto frontier by conditioning on a preference vector $\bm\omega \in \Omega$.
The training pipeline of Envelope SAC is described in \Cref{alg:env_sac} in \Cref{app:algorithms}.

\paragraph{Envelope target.}
We parameterize the critic as a vectorized function $\bm Q_\theta: \mathcal S \times \mathcal A \times \Omega \to \mathbb R^m$, which estimates the expected discounted return for each objective dimension independently. A key challenge in MORL is that the optimal policy for a specific preference query $\bm\omega$ is initially unknown. Standard off-policy updates, which use $a' \sim \pi_\phi(\cdot|s', \bm\omega)$ to update $\bm Q_\theta(\cdot, \cdot, \bm\omega)$, suffer from inefficiency as they ignore potentially superior actions discovered by the policy when conditioned on other preferences.

Given a batch of sampled preferences $\{\bm\omega_k\}_{k=1}^K$, we generate a candidate action set $\mathcal{A}_{\text{cand}} = \{a'_k \sim \pi_\phi(\cdot|s', \bm\omega_k)\}_{k=1}^K$ for the next state $s'$. Instead of restricting the update to the action generated by $\bm\omega_i$, we allow the critic to \emph{"borrow"} the best action from the entire candidate pool. The envelope target $\bm y_i$ for preference $\bm\omega_i$ is computed as:
\begin{equation}
\label{eq:envelope_target}
\begin{aligned}
    \bm y_i &= \bm r + \gamma (1-d) \bm Q_{\bar\theta}(s', a'_{k^*}, \bm\omega_i), \\
    \text{where } k^* &= \operatorname*{arg\,max}_{k \in \{1, \dots, K\}} \left\langle \bm\omega_i, \bm Q_{\bar\theta}(s', a'_k, \bm\omega_i) \right\rangle.
\end{aligned}
\end{equation}
By selecting the index $k^*$ maximizing scalar utility for $\bm\omega_i$, this operator effectively approximates the convex hull of the Pareto front, accelerating convergence by leveraging preference diversity.

\paragraph{Homotopy-based critic update.}
The optimization of the critic presents a two-fold challenge: accurate estimation of the multi-dimensional return vector and maximization of the projected scalar utility. Training solely on scalarized utility leads to information loss regarding individual objectives, while training solely on vector outputs may fail to prioritize the specific preference query $\bm\omega$.

To resolve this, we employ a homotopy-based optimization $\mathcal{L}_Q(\theta)$ \citep{watson1989modern} that linearly interpolates between vector-wise regression and scalar-wise alignment using a dynamic coefficient $\beta$:
\begin{equation}
\label{eq:hybrid_loss}
\begin{split}
    \mathcal{L}_Q(\theta) = \mathbb E_{\mathcal{B}} \Bigl[ & \beta \cdot \underbrace{\left( \langle \bm\omega, \bm Q_\theta(s,a,\bm\omega) \rangle - \langle \bm\omega, \bm y \rangle \right)^2}_{\mathcal{L}_{\text{util}}} + (1-\beta) \cdot \underbrace{\left\| \bm Q_\theta(s,a,\bm\omega) - \bm y \right\|_2^2}_{\mathcal{L}_{\text{vec}}} \Bigr].
\end{split}
\end{equation}
The coefficient $\beta \in [0, 1]$ serves as a curriculum schedule. In the early training stage ($\beta \to 0$), the optimization prioritizes $\mathcal{L}_{\text{vec}}$ to ensure the critic accurately learns the structure of the objective space. As training progresses ($\beta \to 1$), the focus shifts to $\mathcal{L}_{\text{util}}$ to refine the policy alignment with the preferences, ensuring optimality on the Pareto frontier.

\paragraph{Preference-aware policy update.}
The actor $\pi_\phi(a|s, \bm\omega)$ is trained to maximize the expected envelope utility subject to entropy regularization. By conditioning on $\bm\omega$, the policy learns a continuous mapping from preference weights to the optimal action manifold. The policy loss is defined as:
\begin{equation}
\label{eq:policy_loss}
\begin{split}
    \mathcal{L}_\pi(\phi) = \mathbb E_{s \sim \mathcal D, \bm\omega \sim \Omega} \Bigl[ &\alpha \log \pi_\phi(\tilde{a}|s, \bm\omega) - \min_{j=1,2} \left\langle \bm\omega, \bm Q_{\theta_j}(s, \tilde{a}, \bm\omega) \right\rangle \Bigr],
\end{split}
\end{equation}
where $\tilde{a}$ is sampled via the reparameterization trick, and the minimum over two critics ($j=1,2$) is used to mitigate value overestimation \citep{haarnoja2018soft}. This formulation ensures the agent not only adheres to the preference query but also maintains sufficient exploration to discover diverse molecular structures.

\raggedbottom
\section{Experiments}
\label{sec:exp}
To comprehensively validate the effectiveness of our proposed \themodel, we conduct comprehensive experiments on two important scenarios in drug discovery: (1) \emph{binding affinity optimization} (\Cref{subsec:binding}), and (2) \emph{similarity-based rediscovery} (\Cref{subsec:mpo}), both of which can be formulated as multi-objective optimization tasks. We further demonstrate the effectiveness of our model design through ablation experiments and case studies in \Cref{subsec:ablation}.

\paragraph{Experimental Settings}
We adhere to the task settings proposed by \cite{lee2024molecule} for fair comparison. The policy model performs random sampling for the first 4,000 steps to collect experience, and parameter updates after 3,000 steps to ensure a sufficient replay buffer. Every method uses the same evaluation budget of 3,000 completed molecules and docking calls. We report \emph{Novel hit ratio} and \emph{Novel top 5\% Docking Score} for hit discovery \citep{yang21freed}, area under the curve (AUC) of top-$K$ value for rediscovery \citep{gao2022sample}, and chemical-space coverage (\#Circles) \citep{xie2023much}. Please refer to \Cref{app:exp_setting} for experimental settings, evaluation protocols, baselines, and compute profiling.

\begin{table*}[t]
    \caption{\small \textbf{Novel hit ratio (\%) results.} The results are the means and the standard deviations of 3 runs.}
    \centering
    \resizebox{\textwidth}{!}{
    \renewcommand{\tabcolsep}{3mm}
    \begin{tabular}{lcccccc}
    \toprule
        \multirow{2.5}{*}{Method} & \multicolumn{5}{c}{Target protein} \\
    \cmidrule(l{2pt}r{2pt}){2-6}
        & parp1 & fa7 & 5ht1b & braf & jak2 & \\
    \midrule
        REINVENT~\citep{olivecrona2017molecular} & \phantom{0}0.480~\scriptsize($\pm$~0.344) & \phantom{0}0.213~\scriptsize($\pm$~0.081) & \phantom{0}2.453~\scriptsize($\pm$~0.561) & \phantom{0}0.127~\scriptsize($\pm$~0.088) & \phantom{0}0.613~\scriptsize($\pm$~0.167) \\
        Graph GA~\citep{jensen2019graph} & \phantom{0}4.811~\scriptsize($\pm$~1.661) & \phantom{0}0.422~\scriptsize($\pm$~0.193) & \phantom{0}7.011~\scriptsize($\pm$~2.732) & \phantom{0}3.767~\scriptsize($\pm$~1.498) & \phantom{0}5.311~\scriptsize($\pm$~1.667) \\
        MORLD~\citep{jeon20morld} & \phantom{0}0.047~\scriptsize($\pm$~0.050) & \phantom{0}0.007~\scriptsize($\pm$~0.013) & \phantom{0}0.880~\scriptsize($\pm$~0.735) & \phantom{0}0.047~\scriptsize($\pm$~0.040) & \phantom{0}0.227~\scriptsize($\pm$~0.118) \\
        HierVAE~\citep{jin2020hierarchical} & \phantom{0}0.553~\scriptsize($\pm$~0.214) & \phantom{0}0.007~\scriptsize($\pm$~0.013) & \phantom{0}0.507~\scriptsize($\pm$~0.278) & \phantom{0}0.207~\scriptsize($\pm$~0.220) & \phantom{0}0.227~\scriptsize($\pm$~0.127) \\
        RationaleRL~\citep{jin2020multi} & \phantom{0}4.267~\scriptsize($\pm$~0.450) & \phantom{0}0.900~\scriptsize($\pm$~0.098) & \phantom{0}2.967~\scriptsize($\pm$~0.307) & \phantom{0}0.000~\scriptsize($\pm$~0.000) & \phantom{0}2.967~\scriptsize($\pm$~0.196) \\
        FREED~\citep{yang21freed} & \phantom{0}4.627~\scriptsize($\pm$~0.727) & \phantom{0}1.332~\scriptsize($\pm$~0.113) & 16.767~\scriptsize($\pm$~0.897) & \phantom{0}2.940~\scriptsize($\pm$~0.359) & \phantom{0}5.800~\scriptsize($\pm$~0.295) \\
        PS-VAE~\citep{kong2022molecule} & \phantom{0}1.644~\scriptsize($\pm$~0.389) & \phantom{0}0.478~\scriptsize($\pm$~0.140) & 12.622~\scriptsize($\pm$~1.437) & \phantom{0}0.367~\scriptsize($\pm$~0.047) & \phantom{0}4.178~\scriptsize($\pm$~0.933) \\
        MOOD~\citep{lee2023MOOD} & \phantom{0}7.017~\scriptsize($\pm$~0.428) & \phantom{0}0.733~\scriptsize($\pm$~0.141) & 18.673~\scriptsize($\pm$~0.423) & \phantom{0}5.240~\scriptsize($\pm$~0.285) & \phantom{0}9.200~\scriptsize($\pm$~0.524) \\
        GDSS~\citep{jo22GDSS} & \phantom{0}1.933~\scriptsize($\pm$~0.208) & \phantom{0}0.368~\scriptsize($\pm$~0.103) & \phantom{0}4.667~\scriptsize($\pm$~0.306) & \phantom{0}0.167~\scriptsize($\pm$~0.134) & \phantom{0}1.167~\scriptsize($\pm$~0.281) \\
        RetMol~\citep{wang2022retrieval} & \phantom{0}0.011~\scriptsize($\pm$~0.016) & \phantom{0}0.000~\scriptsize($\pm$~0.000) & \phantom{0}0.033~\scriptsize($\pm$~0.027) & \phantom{0}0.000~\scriptsize($\pm$~0.000) & \phantom{0}0.011~\scriptsize($\pm$~0.016) \\
    
        GEAM~\citep{lee2024molecule} & 40.635~\scriptsize($\pm$~0.763) & 17.903~\scriptsize($\pm$~1.220) & 36.667~\scriptsize($\pm$~1.929) & 
        27.887~\scriptsize($\pm$~1.584) & 42.025~\scriptsize($\pm$~1.187) \\
        
        MOLLEO~\citep{wang2025efficient} & 35.725~\scriptsize($\pm$~0.825) & 20.651~\scriptsize($\pm$~1.573) & 32.489~\scriptsize($\pm$~0.533) & 
        25.112~\scriptsize($\pm$~1.837) & 39.120~\scriptsize($\pm$~1.025) \\
        
        \midrule
        \themodel (ours) & \textbf{42.617}~\scriptsize($\pm$~0.733) & \textbf{24.580}~\scriptsize($\pm$~0.652) & \textbf{42.149}~\scriptsize($\pm$~0.125) & 
        \textbf{30.767}~\scriptsize($\pm$~0.974) & \textbf{44.864}~\scriptsize($\pm$~0.982) \\
    \bottomrule
    \end{tabular}}
    \label{tab:novel_hit}
\end{table*}

\begin{table*}[t!]
    \caption{\small \textbf{Novel top 5\% docking score (kcal/mol) results.} The results are the means and the standard deviations of 3 runs. The best results are highlighted in bold.}
    \centering
    \resizebox{\textwidth}{!}{
    \renewcommand{\tabcolsep}{3mm}
    \begin{tabular}{lccccc}
    \toprule
        \multirow{2.5}{*}{Method} & \multicolumn{5}{c}{Target protein} \\
    \cmidrule(l{2pt}r{2pt}){2-6}
        & parp1 & fa7 & 5ht1b & braf & jak2 \\
    \midrule
        REINVENT~\citep{olivecrona2017molecular} & \phantom{0}-8.702~\scriptsize($\pm$~0.523) & -7.205~\scriptsize($\pm$~0.264) & \phantom{0}-8.770~\scriptsize($\pm$~0.316) & \phantom{0}-8.392~\scriptsize($\pm$~0.400) & \phantom{0}-8.165~\scriptsize($\pm$~0.277) \\
        
        Graph GA~\citep{jensen2019graph} & -10.949~\scriptsize($\pm$~0.532) & -7.365~\scriptsize($\pm$~0.326) & -10.422~\scriptsize($\pm$~0.670) & -10.789~\scriptsize($\pm$~0.341) & -10.167~\scriptsize($\pm$~0.576) \\
        MORLD~\citep{jeon20morld} & \phantom{0}-7.532~\scriptsize($\pm$~0.260) & -6.263~\scriptsize($\pm$~0.165) & \phantom{0}-7.869~\scriptsize($\pm$~0.650) & \phantom{0}-8.040~\scriptsize($\pm$~0.337) & \phantom{0}-7.816~\scriptsize($\pm$~0.133) \\
        HierVAE~\citep{jin2020hierarchical} & \phantom{0}-9.487~\scriptsize($\pm$~0.278) & -6.812~\scriptsize($\pm$~0.274) & \phantom{0}-8.081~\scriptsize($\pm$~0.252) & \phantom{0}-8.978~\scriptsize($\pm$~0.525) & \phantom{0}-8.285~\scriptsize($\pm$~0.370) \\
        RationaleRL~\citep{jin2020multi} & -10.663~\scriptsize($\pm$~0.086) & -8.129~\scriptsize($\pm$~0.048) & \phantom{0}-9.005~\scriptsize($\pm$~0.155) & \emph{No hit found} & \phantom{0}-9.398~\scriptsize($\pm$~0.076) \\
        FREED~\citep{yang21freed} & -10.579~\scriptsize($\pm$~0.104) & -8.378~\scriptsize($\pm$~0.044) & -10.714~\scriptsize($\pm$~0.183) & -10.561~\scriptsize($\pm$~0.080) & \phantom{0}-9.735~\scriptsize($\pm$~0.022) \\
        PS-VAE~\citep{kong2022molecule} & \phantom{0}-9.978~\scriptsize($\pm$~0.091) & -8.028~\scriptsize($\pm$~0.050) & \phantom{0}-9.887~\scriptsize($\pm$~0.115) & \phantom{0}-9.637~\scriptsize($\pm$~0.049) & \phantom{0}-9.464~\scriptsize($\pm$~0.129) \\
        MOOD~\citep{lee2023MOOD} & -10.865~\scriptsize($\pm$~0.113) & -8.160~\scriptsize($\pm$~0.071) & -11.145~\scriptsize($\pm$~0.042) & -11.063~\scriptsize($\pm$~0.034) & -10.147~\scriptsize($\pm$~0.060) \\
        GDSS~\citep{jo22GDSS} & \phantom{0}-9.967~\scriptsize($\pm$~0.028) & -7.775~\scriptsize($\pm$~0.039) & \phantom{0}-9.459~\scriptsize($\pm$~0.101) & \phantom{0}-9.224~\scriptsize($\pm$~0.068) & \phantom{0}-8.926~\scriptsize($\pm$~0.089) \\
        RetMol~\citep{wang2022retrieval} & \phantom{0}-8.590~\scriptsize($\pm$~0.475) & -5.448~\scriptsize($\pm$~0.688) & \phantom{0}-6.980~\scriptsize($\pm$~0.740) & \phantom{0}-8.811~\scriptsize($\pm$~0.574) & \phantom{0}-7.133~\scriptsize($\pm$~0.242) \\
        GEAM~\citep{lee2024molecule} & -12.536~\scriptsize($\pm$~0.115) & -9.611~\scriptsize($\pm$~0.019) & -12.385~\scriptsize($\pm$~0.059) & 
        -12.165~\scriptsize($\pm$~0.092) & -11.822~\scriptsize($\pm$~0.035) \\
        MOLLEO~\citep{wang2025efficient} & -11.981~\scriptsize($\pm$~0.125) & -8.921~\scriptsize($\pm$~0.008) & -12.588~\scriptsize($\pm$~0.039) & 
        -11.365~\scriptsize($\pm$~0.117) & -10.988~\scriptsize($\pm$~0.019) \\
        \midrule
        \themodel (ours) & \textbf{-12.930}~\scriptsize($\pm$~0.055) & \textbf{-10.189}~\scriptsize($\pm$~0.023) & \textbf{-13.028}~\scriptsize($\pm$~0.097) & 
        \textbf{-12.497}~\scriptsize($\pm$~0.087) & \textbf{-12.011}~\scriptsize($\pm$~0.015) \\
    \bottomrule
    \end{tabular}}
    \label{tab:novel_top5}
\end{table*}

\begin{table*}[t!]
    \caption{\small \textbf{\#Circles of generated hit molecules.} The \#Circles threshold is set to 0.75. The results are the means and the standard deviations of 3 runs. The best results are highlighted in bold.}
    \centering
    \resizebox{0.95\textwidth}{!}{
    \renewcommand{\tabcolsep}{4mm}
    \begin{tabular}{lccccc}
    \toprule
        \multirow{2.5}{*}{Method} & \multicolumn{5}{c}{Target protein} \\
    \cmidrule(l{2pt}r{2pt}){2-6}
        & parp1 & fa7 & 5ht1b & braf & jak2 \\
    \midrule
        REINVENT~\citep{olivecrona2017molecular} & \phantom{0}44.2~\scriptsize($\pm$~15.5) & 23.2~\scriptsize($\pm$~6.6) & 138.8~\scriptsize($\pm$~19.4) & 18.0~\scriptsize($\pm$~2.1) & \phantom{0}59.6~\scriptsize($\pm$~\phantom{0}8.1) \\
        MORLD~\citep{jeon20morld} & \phantom{0}\phantom{0}1.4~\scriptsize($\pm$~\phantom{0}1.5) & \phantom{0}0.2~\scriptsize($\pm$~0.4) & \phantom{0}22.2~\scriptsize($\pm$~16.1) & \phantom{0}1.4~\scriptsize($\pm$~1.2) & \phantom{0}\phantom{0}6.6~\scriptsize($\pm$~\phantom{0}3.7) \\
        HierVAE~\citep{jin2020hierarchical} & \phantom{0}\phantom{0}4.8~\scriptsize($\pm$~\phantom{0}1.6) & \phantom{0}0.8~\scriptsize($\pm$~0.7) & \phantom{0}\phantom{0}5.8~\scriptsize($\pm$~\phantom{0}1.0) & \phantom{0}3.6~\scriptsize($\pm$~1.4) & \phantom{0}\phantom{0}4.8~\scriptsize($\pm$~\phantom{0}0.7) \\
        RationaleRL~\citep{jin2020multi} & \phantom{0}61.3~\scriptsize($\pm$~\phantom{0}1.2) & \phantom{0}2.0~\scriptsize($\pm$~0.0) & \textbf{312.7}~\scriptsize($\pm$~\phantom{0}6.3) & \phantom{0}1.0~\scriptsize($\pm$~0.0) & \textbf{199.3}~\scriptsize($\pm$~\phantom{0}7.1) \\
        FREED~\citep{yang21freed} & \phantom{0}34.8~\scriptsize($\pm$~\phantom{0}4.9) & 21.2~\scriptsize($\pm$~4.0) & \phantom{0}88.2~\scriptsize($\pm$~13.4) & 34.4~\scriptsize($\pm$~8.2) & \phantom{0}59.6~\scriptsize($\pm$~\phantom{0}8.2) \\
        PS-VAE~\citep{kong2022molecule} & \phantom{0}38.0~\scriptsize($\pm$~\phantom{0}6.4) & 18.0~\scriptsize($\pm$~5.9) & 180.7~\scriptsize($\pm$~11.6) & 16.0~\scriptsize($\pm$~0.8) & \phantom{0}83.7~\scriptsize($\pm$~11.9) \\
        MOOD~\citep{lee2023MOOD} & \phantom{0}86.4~\scriptsize($\pm$~11.2) & 19.2~\scriptsize($\pm$~4.0) & 144.4~\scriptsize($\pm$~15.1) & 50.8~\scriptsize($\pm$~3.8) & \phantom{0}81.8~\scriptsize($\pm$~\phantom{0}5.7) \\
        GEAM~\citep{lee2024molecule} & 123.0~\scriptsize($\pm$~\phantom{0}7.8) & 79.0~\scriptsize($\pm$~9.2) & 144.3~\scriptsize($\pm$~\phantom{0}8.6) & 84.7~\scriptsize($\pm$~8.6) & 118.3~\scriptsize($\pm$~\phantom{0}0.9) \\
        MOLLEO~\citep{wang2025efficient} & 74.2~\scriptsize($\pm$~\phantom{0}2.5) & 56.1~\scriptsize($\pm$~3.2) & 125.3~\scriptsize($\pm$~\phantom{0}7.9) & 68.1~\scriptsize($\pm$~5.4) & 95.2~\scriptsize($\pm$~\phantom{0}3.3) \\
        \midrule
        \themodel (ours) & \textbf{136.7}~\scriptsize($\pm$~4.8) & \textbf{97.5}~\scriptsize($\pm$~3.3) & 173.2 ~\scriptsize($\pm$~7.1) & \textbf{103.2}~\scriptsize($\pm$~5.1) & 132.3~\scriptsize($\pm$~0.9) \\
    \bottomrule
    \end{tabular}}
    \label{tab:ncircles}
\end{table*}

\subsection{Optimization of Binding Affinity}
\label{subsec:binding}

For binding affinity optimization, we aim to design novel, drug-like, and synthesizable binders to the specific target. Therefore, we validate \themodel on five protein targets, \textbf{parp1}, \textbf{fa7}, \textbf{5ht1b}, \textbf{braf}, and \textbf{jak2}, under the quantitative estimate of drug-likeness (QED) \citep{bickerton2012quantifying}, synthetic accessibility (SA) \citep{ertl2009estimation}, and novelty constraints following \cite{lee2023MOOD}.

\begin{table*}[h!]
    \caption{\small \textbf{GuacaMol MPO AUC Top-100 results.} The results are the means of 3 runs. We highlight the best- and the second-performing results in \textbf{boldface} and \underline{underlined}, respectively. }
    \centering
    \resizebox{\textwidth}{!}{
    \renewcommand{\tabcolsep}{1.1mm}
    \begin{tabular}{lccccccccc}
    \toprule
        \multirow{2.5}{*}{Method} & \multicolumn{7}{c}{Benchmark} & \multirow{2.5}{*}{Average} \\
    \cmidrule(l{2pt}r{2pt}){2-8}
        & Amlodipine & Fexofenadine & Osimertinib & Perindopril & Ranolazine & Sitagliptin & Zaleplon & \\
    \midrule
        REINVENT ~\citep{olivecrona2017molecular} & 0.608 & 0.752 & 0.806 & 0.511 & 0.719 & 0.006 & 0.325 &   0.532 \\
        MolDQN ~\citep{zhou2019optimization} & 0.230 & 0.431 & 0.636 & 0.125 & 0.018 & 0.000 & 0.002 &   0.206 \\
        Graph GA ~\citep{jensen2019graph} & 0.622 & 0.731 & 0.799 & 0.503 & 0.670 & 0.330 & 0.305 &   0.566 \\
        Graph MCTS ~\citep{jensen2019graph} & 0.385 & 0.522 & 0.655 & 0.219 & 0.694 & 0.117 & 0.165 &   0.394 \\
        DoG-Gen ~\citep{bradshaw2020barking} & 0.489 & 0.640 & 0.706 & 0.422 & 0.601 & 0.015 & 0.073 &   0.421 \\
        GPBO ~\citep{tripp2021fresh} & 0.538 & 0.685 & 0.750 & 0.460 & 0.694 & 0.117 & 0.165 &   0.487 \\
        STONED ~\citep{nigam2021beyond} & 0.593 & 0.777 & 0.799 & 0.472 & 0.738 & 0.351 & 0.307 &   0.577 \\
        SynNet ~\citep{gao2021amortized} & 0.533 & 0.720 & 0.759 & 0.512 & 0.690 & 0.007 & 0.223 &   0.492 \\
        DST ~\citep{fu2021differentiable} & 0.469 & 0.690 & 0.742 & 0.425 & 0.579 & 0.017 & 0.089 &   0.430 \\
        GEAM ~\citep{lee2024molecule} & \underline{0.627} & \underline{0.786} & 0.817 & 0.514 & 0.714 & 0.408 & 0.386 &   \underline{0.607} \\
        Augmented Memory ~\citep{guo2024augmented} & 0.607 & 0.657 & \underline{0.818} & \underline{0.518} & \underline{0.725} & 0.398 & 0.355 &   0.582 \\
        MOLLEO (MoleculeSTM) ~\citep{wang2025efficient} & 0.617 & 0.765 & 0.799 & 0.516 & 0.675 & \underline{0.415} & \underline{0.392} &   0.597 \\
        \midrule
        \themodel (Ours) & \textbf{0.671} & \textbf{0.842} & \textbf{0.845} & \textbf{0.533} & \textbf{0.748} & \textbf{0.489} & \textbf{0.418} &  \textbf{0.649} \\
    \bottomrule
    \end{tabular}}
    \label{tab:mpo}
\end{table*}

\textbf{Performance comparison.} Empirical results are summarized in \Cref{tab:novel_hit,tab:novel_top5,tab:ncircles}.
Our systematic study suggests the following trends:  \emph{(i) \textbf{\themodel} achieves state-of-the-art performance across all targets}, highlighting its ability to generate novel, drug-like, synthesizable compounds with strong binding affinity. In particular, for challenging targets such as \textit{fa7}, our method attains a novel hit ratio above 24\%, surpassing GEAM (17\%) and far outperforming most competing methods, many of which approach zero. \emph{(ii) Fragment-based reinforcement learning methods are especially effective at exploring high-reward regions.} FREED, MOOD, GEAM, and our \themodel all adopt fragment-based RL strategies and consistently produce lower docking scores. This advantage arises from a well-defined action space: fragment-level operations are limited yet high-quality, enabling more efficient exploration than token-based SMILES generation. \emph{(iii) \textbf{\themodel} achieves higher \#Circle while maintaining a strong hit ratio}, indicating that the generated molecules are not only effective but also diverse. This reflects the benefit of explorative preferences, which encourage RL to explore a wider range of action sequences and thereby cover chemical space more comprehensively. By contrast, RationaleRL achieves relatively high diversity on \textit{5ht1b} and \textit{jak2}, but its overly aggressive exploration leads to very low hit discovery rates.

\begin{wrapfigure}{r}{.4\textwidth}
\vspace{-4mm}
\centerline{
\begin{tabular}{c}
\hspace*{-2mm}\includegraphics[width=.3\textwidth,height=!]{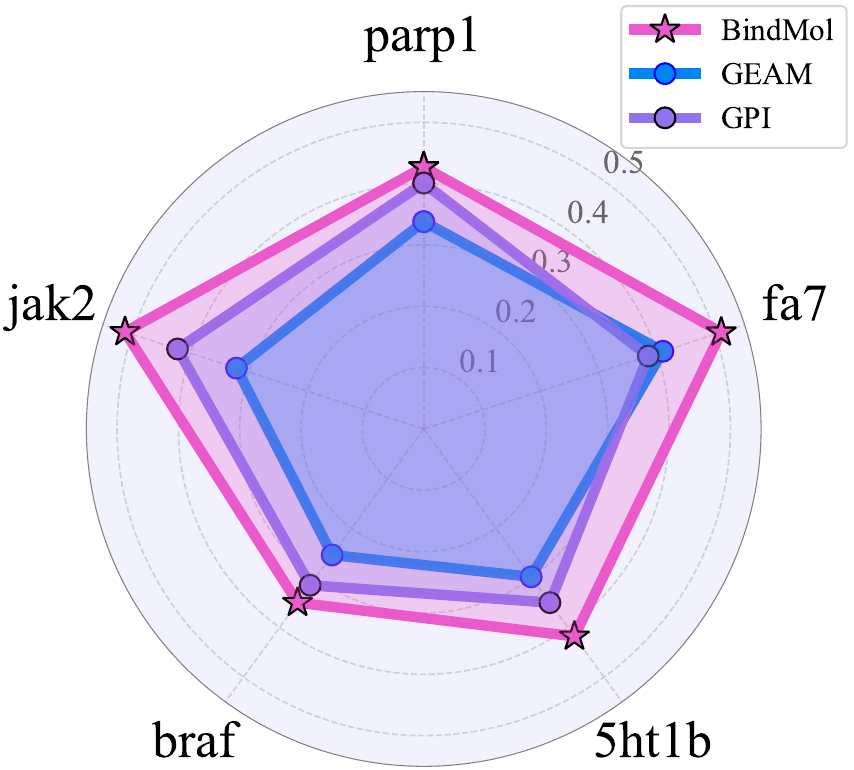}
\end{tabular}}
\vspace{-1mm}
\caption{The Hypervolume comparison of \themodel.}
\vspace{-3mm}
\label{fig:radar}
\end{wrapfigure}

\textbf{Multi-residue exploration.} We evaluate the ability of \themodel to explore diverse binding modes using the \textbf{Hypervolume} metric, which quantifies the dominated volume of the Pareto front with respect to a preference point. As multi-objective hotspot optimization typically involves three to five objectives, many existing methods become computationally prohibitive. We therefore compare against GEAM \cite{lee2024molecule}, a strong baseline for binding tasks, and GPI \citep{alegre2023sample}, a multi-objective reinforcement learning method scalable to 3--5 objectives.
As shown in \Cref{fig:radar}, multi-objective RL methods achieve broader Pareto-front coverage than GEAM, reflecting improved balance across competing interaction hotspots. Our method consistently attains the most extensive coverage across all five objectives, spanning multiple trade-off directions. This uniform Pareto exploration indicates that \themodel avoids collapsing to a single optimization mode and instead generates molecules with more diverse binding modes.

\begin{table}[!htbp]
    \caption{\small \textbf{Direct binding-mode and matched MORL evaluation.} IFP statistics are computed from full-pocket PLIP residue--interaction-type fingerprints; GEAM is the strongest baseline on all three statistics. Recovery denotes the percentage of generated hits with IFP Jaccard similarity $\geq 0.4$ to the native crystal ligand. MORL results average the five targets; complete per-target results and protocols are in \Cref{app:interaction_eval}.}
    \label{tab:direct_mode}
    \centering
    \small
    \setlength{\tabcolsep}{3pt}
    \textbf{(a) Interaction-pattern diversity}\par\smallskip
    \begin{tabular*}{\textwidth}{@{\extracolsep{\fill}}llccccc@{}}
    \toprule
    Metric & Method & PARP1 & FA7 & 5HT1B & BRAF & JAK2\\
    \midrule
    \multirow{2}{*}{IFP clusters} & GEAM & 138.67 & 61.67 & 8.33 & 20.33 & 47.33\\
    & \themodel & \textbf{192.00} & \textbf{94.33} & \textbf{11.67} & \textbf{35.00} & \textbf{77.33}\\
    \midrule
    \multirow{2}{*}{Entropy $H$} & GEAM & 2.93 & 1.85 & 0.24 & 0.68 & 1.52\\
    & \themodel & \textbf{3.45} & \textbf{2.37} & \textbf{0.32} & \textbf{1.11} & \textbf{2.31}\\
    \midrule
    \multirow{2}{*}{Effective modes} & GEAM & 18.73 & 6.36 & 1.27 & 1.97 & 4.57\\
    & \themodel & \textbf{31.40} & \textbf{10.71} & \textbf{1.37} & \textbf{3.03} & \textbf{10.10}\\
    \bottomrule
    \end{tabular*}
    \par\vspace{2mm}
    \begin{minipage}[t]{0.57\textwidth}
    \centering\small
    \textbf{(b) Crystallographic-mode recovery (\%)}\par\smallskip
    \begin{tabular*}{\linewidth}{@{\extracolsep{\fill}}lrrr@{}}
    \toprule
    Mode & \themodel & GEAM & FREED\\
    \midrule
    BRAF DFG-in / I & \textbf{48.9} & 20.3 & 11.7\\
    BRAF DFG-out / II & \textbf{12.1} & 6.4 & 0.0\\
    JAK2 type-I & \textbf{1.3} & 0.0 & 0.0\\
    JAK2 type-II & \textbf{11.5} & 5.7 & 2.8\\
    \bottomrule
    \end{tabular*}
    \end{minipage}\hfill
    \begin{minipage}[t]{0.39\textwidth}
    \centering\small
    \textbf{(c) Matched MORL averages}\par\smallskip
    \begin{tabular*}{\linewidth}{@{\extracolsep{\fill}}lrr@{}}
    \toprule
    Method & Hit (\%) & DS\\
    \midrule
    GPI & 32.15 & $-10.60$\\
    C-MORL & 32.35 & $-10.63$\\
    PreCo & 32.87 & $-10.73$\\
    \themodel & \textbf{37.01} & $\mathbf{-12.13}$\\
    \bottomrule
    \end{tabular*}
    \end{minipage}
\end{table}

\textbf{Direct binding-mode diversity.} Unlike \#Circles and Hypervolume, which measure ligand-space and objective-space coverage, respectively, the PLIP-IFP metrics in \Cref{tab:direct_mode} directly compare docking-predicted interaction patterns. \themodel yields more clusters and more balanced mode populations than GEAM on every target. We additionally select BRAF and JAK2 because they provide well-established, within-target pairs of type-I/DFG-in and type-II/DFG-out crystal complexes; the other targets lack equally clear paired mode labels. Reference-pair IFP/residue Jaccard similarities are 0.375/0.462 for BRAF and 0.188/0.250 for JAK2, and \themodel recovers all four crystallographic modes more frequently than the baselines.

\textbf{Matched MORL comparison.} We adapt GPI, C-MORL, and PreCo to the same molecular action space, hotspot reward vector, training protocol, docking budget, and three seeds. \Cref{tab:direct_mode}(c) summarizes the gains, with full per-target results in \Cref{tab:morl_full}. Our claim is specific to preference-conditioned molecular exploration with 3--5 hotspot objectives, rather than general-purpose MORL superiority.

\subsection{Optimization of Multi-property Objectives in GuacaMol}
\label{subsec:mpo}

For similarity-based rediscovery, our goal is to generate molecules similar to the target compound and fulfill multiple required property profiles. Following \cite{gao2022sample}, we select the multi-property optimization tasks from the GuacaMol benchmarks \citep{brown2019guacamol} for capability evaluation. 
Notably, unlike prior studies that scalarize multiple objectives into a single target, we treat them as a genuine multi-objective problem to better match our algorithm. To ensure fairness, we use each individual objective score prior to scalarization as the single entry of multi-objective reward.

\textbf{Performance comparison.} Empirical results for MPO tasks are presented in \Cref{tab:mpo} and \Cref{tab:mpo_novel}. It can be observed that \themodel significantly outperforms all baseline models across all MPO tasks, which further demonstrates the effectiveness of the proposed Envelope SAC algorithm. On the two rediscovery tasks of \emph{Fexofenadine} and \emph{Sitagliptin}, it achieves relative improvements of 7.1\% and 17.8\% over the second-best models, respectively. Moreover, on the most challenging \emph{Zaleplon} rediscovery task, \themodel attains a record-breaking score of 0.418. It is worth noting that MOLLEO (MoleculeSTM), a framework that applies genetic algorithms over large-scale molecular corpora with language models, is still outperformed by our method on rediscovery tasks. We exclude other variants of MOLLEO (e.g., those using GPT-4 as the base LLM) from comparison due to their unfair advantage in this setting.

\subsection{Ablation Studies}
\label{subsec:ablation}

\begin{figure*}[t]
    \centering
    \includegraphics[width=\textwidth,height=!]{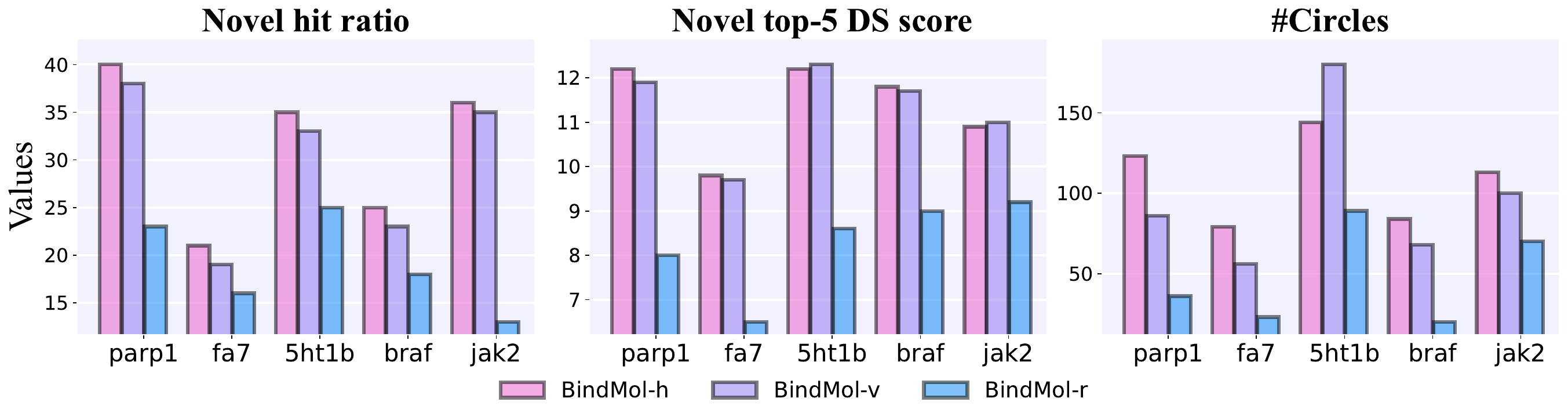}
    \caption{Performance of different ablation variants of \themodel. DS denotes Docking Score.}
    \label{fig:ablation}
    \vspace{-3mm}
\end{figure*}

\begin{figure*}[h!]
    \centering
    \includegraphics[width=\textwidth,height=!]{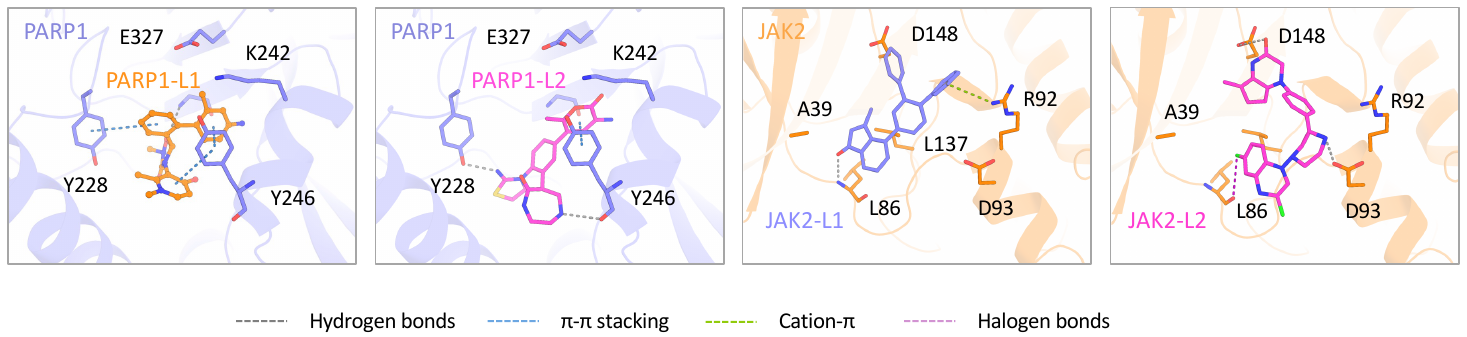}
    \caption{Visualization of representative binding modes on two protein complexes: \emph{parp1} and \emph{jak2}.}
    \label{fig:case}
    \vspace{-3mm}
\end{figure*}

We perform ablation studies to isolate the contributions of the key components in our framework.
Specifically, we remove energy-based fragment scoring and construct the action space from a default fragment library (\emph{\themodel-r}); ablate the hotspot-aware reward and use only target-wise docking scores for optimization (\emph{\themodel-h}); and remove the envelope target, reducing Envelope SAC to standard SAC with vector-valued rewards (\emph{\themodel-v}).
As shown in \Cref{fig:ablation}, removing any component leads to consistent performance degradation in both binding affinity and chemical-space coverage. The energy-based vocabulary most directly improves affinity-oriented search, explaining the larger decline of \emph{\themodel-r} on docking metrics. The hotspot reward serves a different role: removing it reduces BRAF IFP clusters from $35.00\!\pm\!1.41$ to $25.67\!\pm\!1.53$ and effective modes from $3.03\!\pm\!0.08$ to $2.20\!\pm\!0.08$ (\Cref{tab:hotspot_ablation}), directly supporting its contribution to interaction-pattern diversity. The envelope update further improves preference-conditioned exploration, as reflected by \emph{\themodel-v}.

\subsection{Case Studies}

To provide an intuitive understanding of the binding modes exploration of \themodel, we cluster the generated molecules according to the number of protein--ligand interactions across different targets using $k$-means ($k=2$). For each cluster, we select and visualize representative samples to examine their interaction patterns. Specifically, we perform clustering analysis on the \emph{parp1} and \emph{jak2} targets, resulting in four protein--ligand complexes for visualization, as shown in \Cref{fig:case}.

In the \emph{parp1} complexes, the two ligands exhibit distinct binding modes: PARP1-L1, with a polyaromatic scaffold, is dominated by aromatic stacking with Y228 and Y246, whereas PARP1-L2, featuring a heteroaromatic scaffold, relies more heavily on hydrogen bonding with diminished stacking.
In the \emph{jak2} complexes, the differences are also evident. JAK2-L1 adopts a stacking- and cation--$\pi$-driven mode, while JAK2-L2 introduces heteroatoms and halogen substitution that favor additional hydrogen and halogen bonds, reducing dependence on R92.

\section{Conclusion}

In this work, we revisit target-based drug discovery from the perspective of binding-mode diversity and argue that optimizing a single scalar objective is insufficient to capture the complexity of protein--ligand interactions. To this end, we propose \themodel, a multi-objective reinforcement learning framework that explicitly models hotspot-level interactions and their inherent trade-offs. By combining energy-based fragment scoring, biologically grounded vectorized rewards, and the \emph{Envelope SAC} algorithm for preference-aware optimization, \themodel enables efficient exploration of diverse, high-affinity binding modes within the same target. Experiments across multiple targets and benchmarks demonstrate consistent improvements over existing approaches.

\paragraph{Broader impacts.} Given that our application tasks fall within the molecular domain, improper use of methods for tasks such as molecular property prediction may result in significant deviations. This could impact subsequent applications of the molecules in drug development or materials design, especially in predicting properties like toxicity and stability. We recommend further experimental validation of key molecules after using the model to ensure the reliability of the results.

\paragraph{Acknowledgements.} This work is jointly supported by National Science and Technology Major Project (2023ZD0120901) and Zhongguancun Academy (XTS0047, C20250208).

\clearpage
\bibliography{reference}
\bibliographystyle{reference}

\newpage
\appendix
\clearpage
\section{Algorithms}
\label{app:algorithms}

\begin{algorithm}[H]
    \caption{Energy-Based Fragment Scoring}
    \label{alg:frag_score}
    \KwIn{Molecular dataset $\mathcal{D}$, property labels $\mathcal{Y}$, batch size $B$, learning rate $\eta$}
    Initialize energy network $E_\theta$ parametrized by GNN, and empty vocabulary $\mathcal{V} \leftarrow \emptyset$\;
    
    \textcolor{VioletRed}{\tcp{Phase 1: Energy Landscape Learning}}
    \Repeat{Convergence}{
        Sample batch of positive pairs $B_{\text{real}} = \{(G_i, y_i)\}_{i=1}^B$ from $\mathcal{D}$\;
        
        \For{each sample $(G, y)$ in $B_{\text{real}}$}{
            Sample noisy property $\tilde{y} \sim p_{\text{uniform}}(\mathcal{Y})$\;
            
            Generate corrupted graph $\tilde{G}$ by masking random node features: $\mathbf{H}_{\tilde{G}} \leftarrow \mathbf{H}_G \odot \mathbf{M}_{\text{rand}}$\;

            Compute energy for real and noisy pairs:\;
            $E_{\text{real}} \leftarrow E_\theta(G, y)$; \quad $E_{\text{noise}} \leftarrow E_\theta(\tilde{G}, \tilde{y})$\;
            
            Compute InfoNCE loss to shape the landscape:\;
            $\mathcal{L}_{\text{NCE}} \leftarrow \log \left( 1 + \exp(E_{\text{real}} - E_{\text{noise}}) \right)$\;
        }
        
        Update parameters $\theta$ by minimizing the batch loss: $\theta \leftarrow \theta - \eta \nabla_\theta \sum \mathcal{L}_{\text{NCE}}$\;
    }

    \textcolor{VioletRed}{\tcp{Phase 2: Goal-Aware Fragment Extraction}}
    Filter high-activity molecules $\mathcal{D}_{\text{high}} \subset \mathcal{D}$\;
    \For{each molecule $G \in \mathcal{D}_{\text{high}}$}{
        Decompose $G$ into candidate fragments $\{F_1, \dots, F_k\}$ (e.g., via BRICS)\;
        
        \For{each fragment $F_j$}{
            Calculate Importance Score (Energy Gap):\;
            $\mathcal{I}(F_j) \leftarrow E_\theta(G_{\setminus F_j}, y_{\text{high}}) - E_\theta(G, y_{\text{high}})$\;
            
            \If{$\mathcal{I}(F_j) > \tau$}{
                Add fragment to vocabulary: $\mathcal{V} \leftarrow \mathcal{V} \cup \{F_j\}$\;
            }
        }
    }
    \KwOut{Goal-Aware Fragment Vocabulary $\mathcal{V}$}
\end{algorithm}

\begin{algorithm}[H]
    \caption{Envelope Soft Actor-Critic}
    \label{alg:env_sac}
    \KwIn{Preference sampling distribution $\Omega$, batch size $B$, number of sampled preferences $K$}
    Initialize policy $\pi_\phi$, critics $\bm Q_{\theta_1, \theta_2}$, target critics $\bm Q_{\bar{\theta}_1, \bar{\theta}_2}$, and replay buffer $\mathcal{D}$\;
    
    \Repeat{Convergence}{
        Interact with the environment using $\pi_\phi(\cdot|s,\bm\omega_{\text{env}})$, and store transitions $(s,a,\bm r,s',d)$ in $\mathcal{D}$\;
        
        \For{each gradient step}{
            Sample minibatch $\mathcal{B} = \{(s,a,\bm r,s',d)\}$ from $\mathcal{D}$ and preferences $\{\bm\omega_i\}_{i=1}^K \sim \Omega$\;
            
            \textcolor{VioletRed}{\tcp{Critic Update}}
            Generate candidate actions for next state:
            $\mathcal{A}_{\text{cand}} = \{ a'_k \sim \pi_\phi(\cdot|s',\bm\omega_k) \}_{k=1}^K$\;
            
            \For{each preference $\bm\omega_i$}{
                Select envelope action index:
                $k^* \leftarrow \arg\max_{k \in \{1,\dots,K\}} 
                \left\langle \bm\omega_i,\;
                \min_{j=1,2} \bm Q_{\bar{\theta}_j}(s',a'_k,\bm\omega_i) \right\rangle$\;
                
                Compute envelope target vector:\\
                $\bm y_i \leftarrow \bm r + \gamma (1-d)\,
                \min_{j=1,2} \bm Q_{\bar{\theta}_j}(s',a'_{k^*},\bm\omega_i)$\;
            }
            
            Update critics $\theta_{1,2}$ by minimizing the homotopy loss:
            \[
            \mathcal{L}_Q = 
            \mathbb E_{\mathcal{B},\,\bm\omega}
            \Big[
            \beta \big(
            \langle \bm\omega, \bm Q_\theta(s,a,\bm\omega)\rangle -
            \langle \bm\omega, \bm y \rangle
            \big)^2
            + (1-\beta)
            \|\bm Q_\theta(s,a,\bm\omega) - \bm y\|_2^2
            \Big]
            \]
            
            \textcolor{VioletRed}{\tcp{Policy Update}}
            Sample actions $\tilde a \sim \pi_\phi(\cdot|s,\bm\omega)$ using reparameterization\;
            
            Update policy $\phi$ by minimizing:
            $
            \mathcal{L}_\pi =
            \mathbb E_{\mathcal{B},\,\bm\omega}
            \Big[
            \alpha \log \pi_\phi(\tilde a|s,\bm\omega)
            -
            \min_{j=1,2}
            \langle \bm\omega,
            \bm Q_{\theta_j}(s,\tilde a,\bm\omega)
            \rangle
            \Big]
            $
            
            Update target critics:
            $\bar{\theta}_j \leftarrow \rho \bar{\theta}_j + (1-\rho)\theta_j$ for $j=1,2$\;
        }
    }
\end{algorithm}

\section{Experimental Settings}
\label{app:exp_setting}
\subsection{Computing Infrastructures}\label{app:infra}
\paragraph{Software infrastructures.} All of the experiments are implemented in Python 3.10, with the following
supporting libraries: PyTorch 2.1.0 \citep{paszke2019pytorch}, PyG 2.0.4 \citep{fey2019fast}, DGL 2.4.0 \citep{wang2019deep}, Gym 0.21.0 \citep{brockman2016gym}, Open Babel 3.1.1 \citep{boyle2011openbabel}, PLIP 2.3.1 \citep{salentin2015plip}, and RDKit 2022.03.1 \cite{landrum2016rdkit}.
\paragraph{Hardware infrastructures.} We conduct all experiments on a computer server with 8 NVIDIA A800 GPUs and 2 Intel Xeon Gold 6348 CPUs (with 56 cores and 112 threads).

\subsection{Implementation Details}

Following \cite{yang21freed,gao2022sample,lee2024molecule}, we base our experimental setup on the ZINC250k dataset \citep{irwin2012zinc}, using the standard split for training and testing. For our generative model, the training process begins with an initial fragment vocabulary of 300 fragments, which is dynamically expanded to a maximum size of 1000 by adding up to 50 new fragments per cycle. Any fragments that cause RDKit sanitization errors are filtered out.

The fragment-based generator employs Envelope SAC for training, where episode starts from a benzene ring scaffold with three attachment points (\emph{ortho-}, \emph{meta-}, \emph{para-}). An episode terminates if the heavy atom count of molecule exceeds 40. To ensure sufficient exploration, the policy model samples molecules randomly for the first 4,000 steps to populate the replay buffer, while parameter updates commence after 3,000 steps. For fair evaluation, molecules generated during this initial exploration phase are included in the final results. The fragment modification module utilizes a genetic algorithm with a population size of 100 and a mutation rate of 0.1.

All neural networks use a hidden dimension of 64 and are trained with a batch size of 256. Both the value and policy networks are optimized using the Adam optimizer \citep{kingma2014adam} with a learning rate of $1\times10^{-4}$. For exploration in our multi-objective setting, we sample 5 preference vectors per batch. Moreover, we observe that as the number of objectives increases, the training benefits from a higher ratio of off-policy updates (i.e., utilizing more samples from the replay buffer relative to on-policy samples). We hypothesize that in high-dimensional objective spaces, the model requires more frequent re-visitation of past action episodes to effectively learn the complex trade-offs and stabilize convergence.

\paragraph{Oracle budget and runtime.} Each target-based method is evaluated with the same budget of 3,000 completed molecules and docking calls. Docking is invoked only after a molecule is completed; the resulting pose is reused by PLIP, so hotspot profiling requires no additional docking call. On the same server, representative training times are 2.3, 9.7, 11.6, 12.4, and 13.2 hours for REINVENT, FREED, MOOD, GEAM, and \themodel, respectively. REINVENT benefits from its lightweight RNN, while \themodel remains comparable to fragment-based baselines. Preference vectors are replicated across the batch for parallel actor/critic evaluation rather than processed in a Python loop. With fixed poses, multi-threaded PLIP processes $10^4$ molecules within minutes, so its overhead is small relative to docking.

\subsection{Evaluation Protocols}
We adopt several well-defined metrics to evaluate our framework:
\begin{itemize}[leftmargin=*, labelsep=0.5em]
    \item For binding affinity optimization, we evaluate the generated molecules following \cite{yang21freed}. A molecule is considered a \emph{hit} if it achieves a docking score lower than the median of experimentally known actives, while also satisfying $ \text{QED} > 0.5 $ and $ \text{SA} < 5 $. To assess affinity optimization, we report the \textbf{\emph{Novel top 5\% Docking Score}} (kcal/mol), which is the average docking score of the best-performing 5\% unique and novel hits. We also measure the \textbf{\emph{Novel hit ratio}}, defined as the fraction of hits that are novel.
    \item For rediscovery tasks, we report the area under the curve (AUC) of top-K average property value versus the number of oracle calls (\textbf{\emph{AUC top-K}}) following \cite{gao2022sample} for fair comparison.
    \item We further report \textbf{\emph{\#Circles}}~\citep{xie2023much} for both of the tasks, which quantifies the coverage of chemical space by the generated hits.
\end{itemize} 

\subsection{Baselines}
\paragraph{Binding affinity tasks.} To comprehensively evaluate the effectiveness of our approach, we select 11 competitive molecular generation models as baselines: REINVENT \citep{olivecrona2017molecular}, Graph GA \citep{jensen2019graph}, MORLD \citep{jeon20morld}, HierVAE \citep{jin2020hierarchical}, RationaleRL \citep{jin2020multi}, FREED \citep{yang21freed}, PS-VAE \citep{kong2022molecule}, MOOD \citep{lee2023MOOD}, GDSS \citep{jo22GDSS}, RetMol \citep{wang2022retrieval}, GEAM \citep{lee2024molecule} and MOLLEO \citep{wang2025efficient}. These baselines span both SMILES- and graph-based modalities, and encompass diverse training paradigms including distribution learning, genetic algorithms, and reinforcement learning. Note that our model is trained from scratch, rather than fine-tuned from a large pretrained molecular generative model. To ensure a fair and controlled comparison under this setting, we do not include approaches that rely heavily on large-scale pretraining \citep{pandey2025pretraining, lee2025genmol}.

\paragraph{Multi-property tasks in GuacaMol.} We select 12 molecular generation models with strong performance on the GuacaMol benchmark as baselines: REINVENT \citep{olivecrona2017molecular}, MolDQN \citep{zhou2019optimization}, Graph GA \citep{jensen2019graph}, Graph MCTS \citep{jensen2019graph}, DoG-Gen \citep{bradshaw2020barking}, GPBO \citep{tripp2021fresh}, STONED \citep{nigam2021beyond}, SynNet \citep{gao2021amortized}, DST \citep{fu2021differentiable}, GEAM \citep{lee2024molecule}, Augmented Memory \citep{guo2024augmented}, and MOLLEO (MoleculeSTM) \citep{wang2025efficient}. These methods also span various molecular modalities and training strategies.

\section{Spatial Clustering of Hotspot Residues}
\label{app:cluster}

Given the non-overlapping residue set $\mathcal V$ defined in \Cref{subsec:reward}, our goal is to further partition these residues into a set of spatially coherent subpockets that capture distinct local interaction patterns within the binding site. The clustering is performed purely based on the three-dimensional geometry of residues, without relying on ligand-specific distance thresholds.
Each residue $v \in \mathcal V$ is represented by a three-dimensional centroid $\mathbf c_v$, computed from its side-chain atoms to emphasize interaction-relevant geometry. Backbone atoms (N, C$_\alpha$, C, O) are excluded whenever side-chain atoms are available; for residues lacking side chains, the centroid is computed over all atoms of the residue. This representation yields a set of residue-level points embedded in $\mathbb{R}^3$.

We construct a pairwise distance matrix $\mathbf D \in \mathbb{R}^{|\mathcal V| \times |\mathcal V|}$, where each entry
\[
D_{uv} = \lVert \mathbf c_u - \mathbf c_v \rVert_2 , \qquad u, v \in \mathcal V,
\]
quantifies the spatial proximity between residues. Based on this distance matrix, we define a Gaussian affinity graph with weights
\[
W_{uv} = \exp\!\left(-\frac{D_{uv}^2}{\sigma^2}\right),
\]
where $\sigma$ controls the spatial scale at which residue interactions are considered local. The diagonal entries of $\mathbf W$ are set to zero.
To identify spatially separated but internally compact residue groups, we apply spectral clustering to the affinity graph. Specifically, we construct the unnormalized graph Laplacian $\mathbf L = \mathbf D_W - \mathbf W$, where $\mathbf D_W$ is the diagonal degree matrix of $\mathbf W$. The eigenvectors corresponding to the $m$ smallest eigenvalues of $\mathbf L$ are extracted and row-normalized to form a low-dimensional spectral embedding of residues in $\mathcal V$.

\begin{figure}[t]
    \centering
    \setlength{\tabcolsep}{6pt}
    \renewcommand{\arraystretch}{1.15}

    \begin{tabular}{ccccc}

    \begin{subfigure}[t]{0.18\textwidth}
        \centering
        \includegraphics[width=\linewidth]{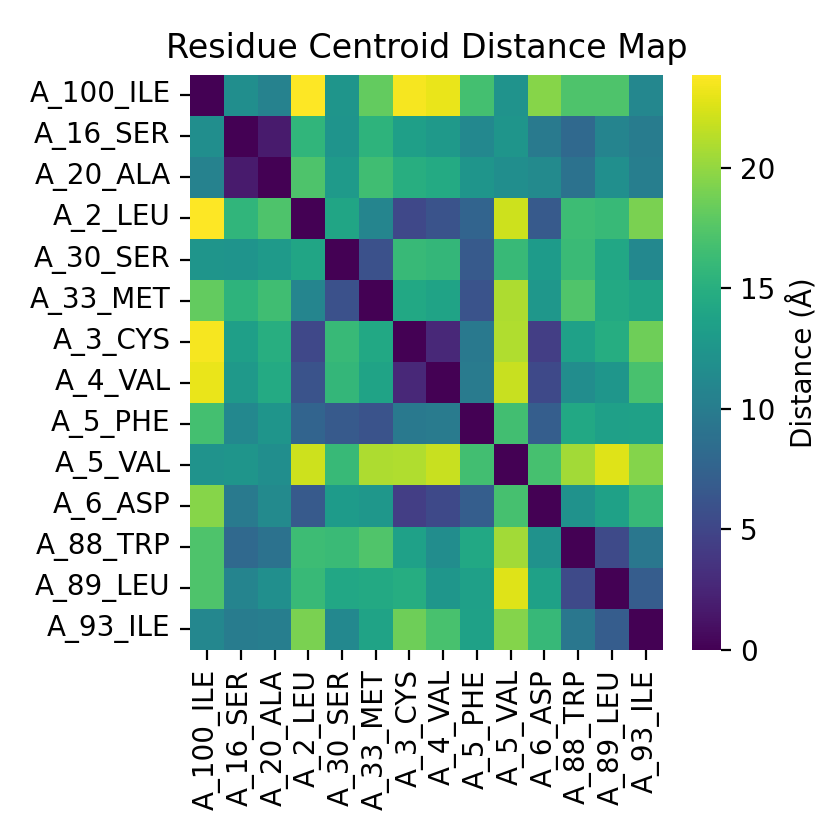}
        \caption{5ht1b}
    \end{subfigure} &
    \begin{subfigure}[t]{0.18\textwidth}
        \centering
        \includegraphics[width=\linewidth]{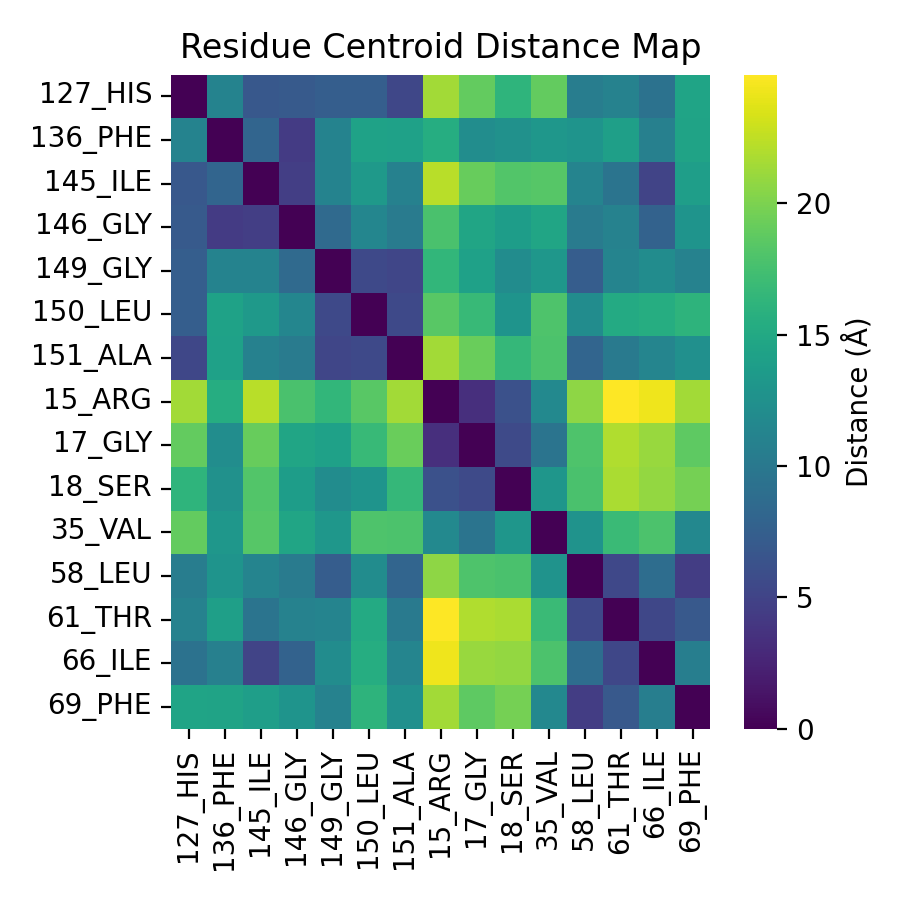}
        \caption{braf}
    \end{subfigure} &
    \begin{subfigure}[t]{0.18\textwidth}
        \centering
        \includegraphics[width=\linewidth]{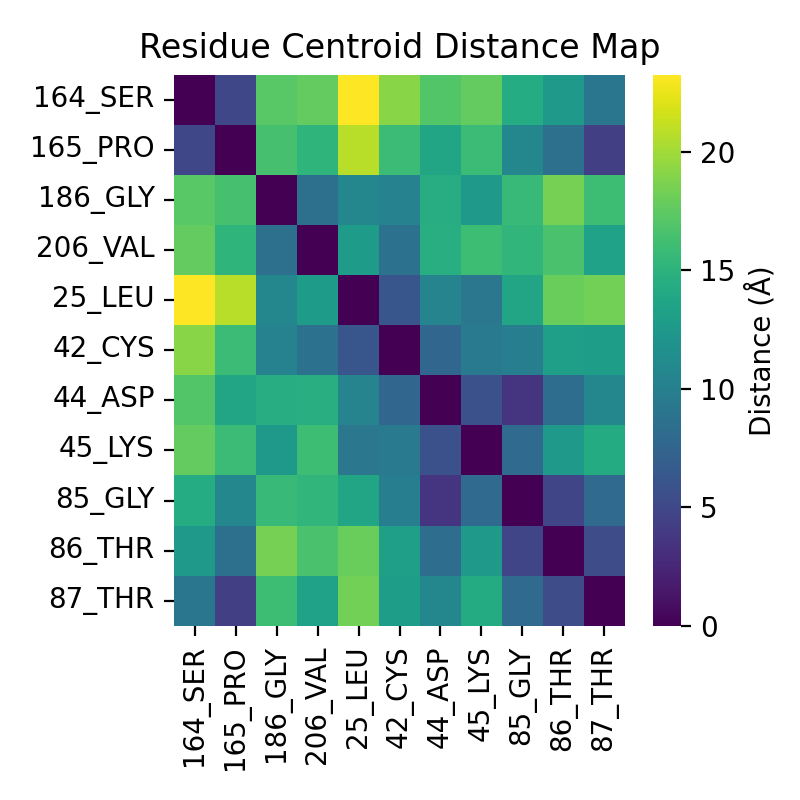}
        \caption{fa7}
    \end{subfigure} &
    \begin{subfigure}[t]{0.18\textwidth}
        \centering
        \includegraphics[width=\linewidth]{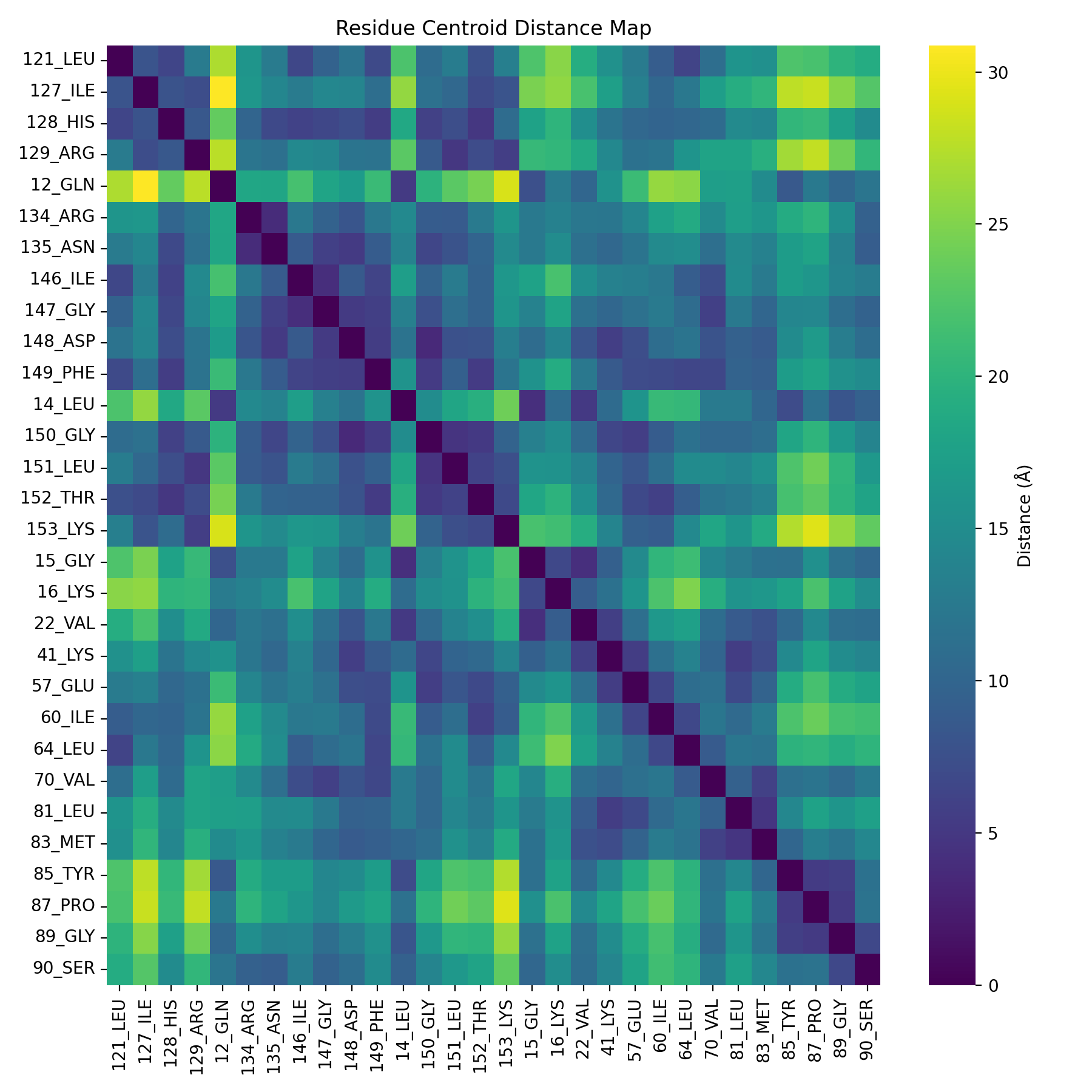}
        \caption{jak2}
    \end{subfigure} &
    \begin{subfigure}[t]{0.18\textwidth}
        \centering
        \includegraphics[width=\linewidth]{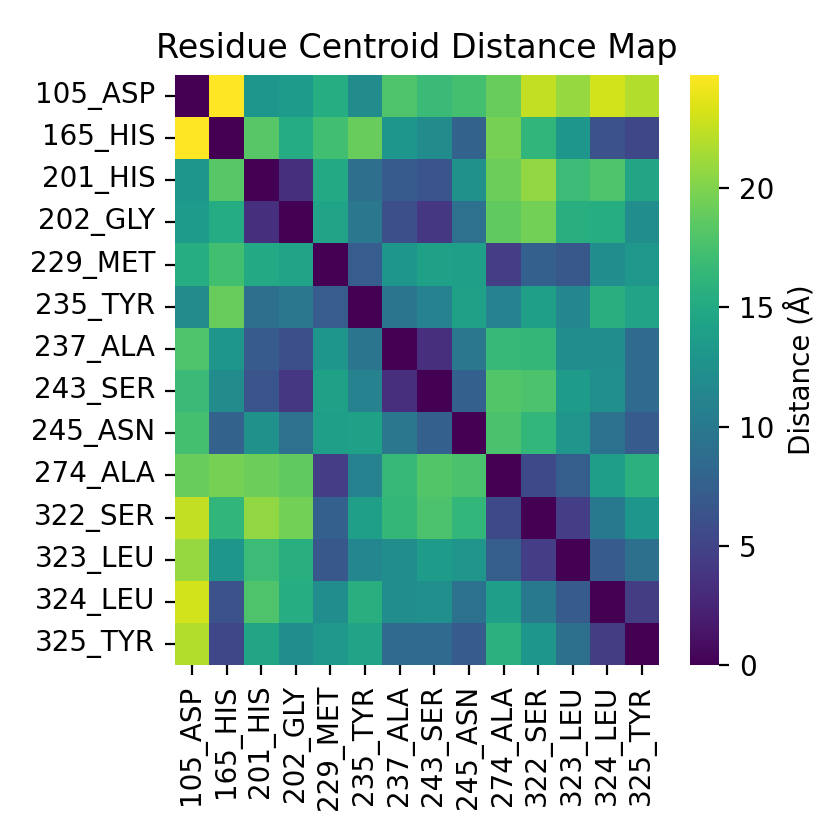}
        \caption{parp1}
    \end{subfigure}
    \\

    \begin{subfigure}[t]{0.18\textwidth}
        \centering
        \includegraphics[width=\linewidth]{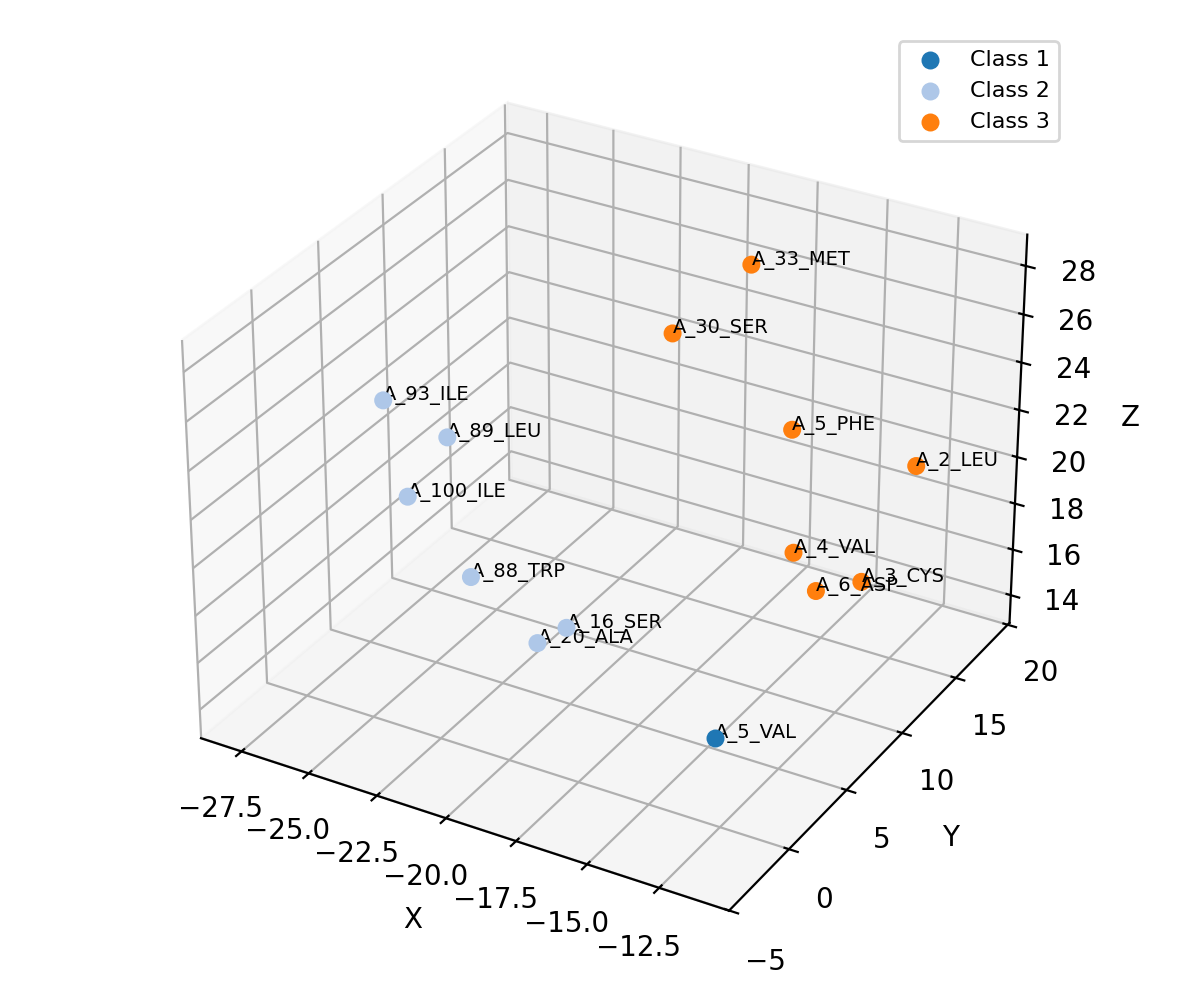}
        \caption{5ht1b}
    \end{subfigure} &
    \begin{subfigure}[t]{0.18\textwidth}
        \centering
        \includegraphics[width=\linewidth]{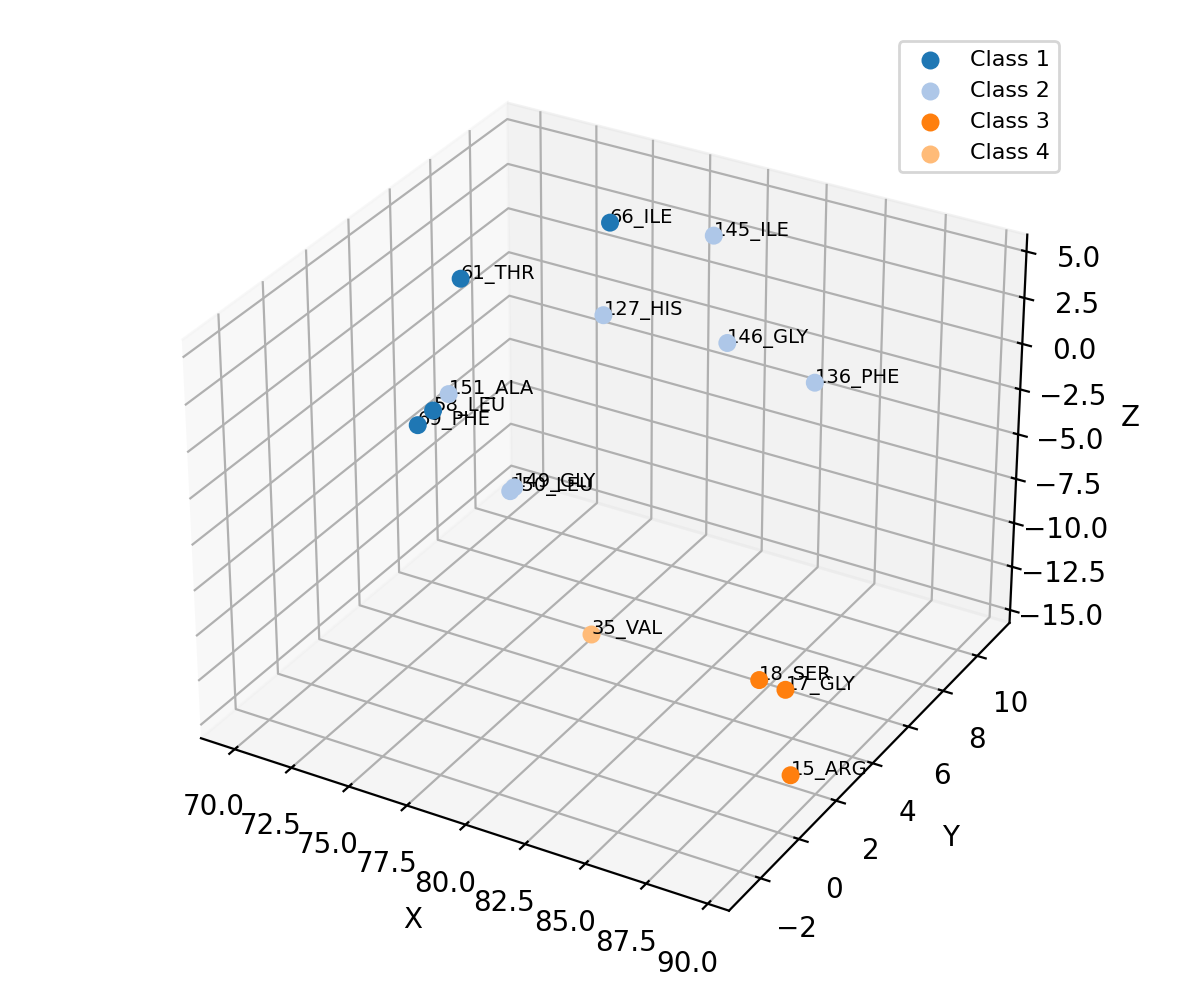}
        \caption{braf}
    \end{subfigure} &
    \begin{subfigure}[t]{0.18\textwidth}
        \centering
        \includegraphics[width=\linewidth]{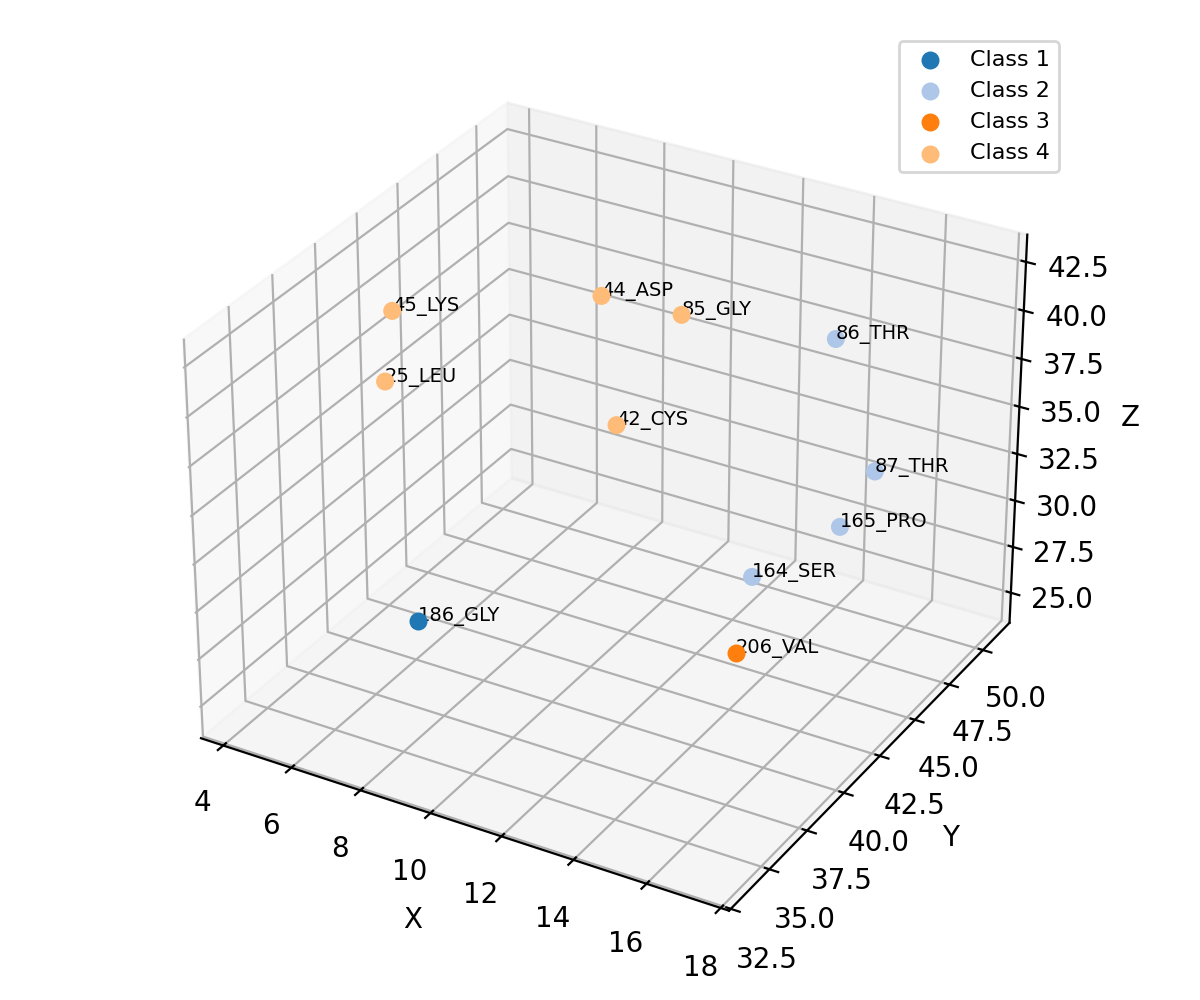}
        \caption{fa7}
    \end{subfigure} &
    \begin{subfigure}[t]{0.18\textwidth}
        \centering
        \includegraphics[width=\linewidth]{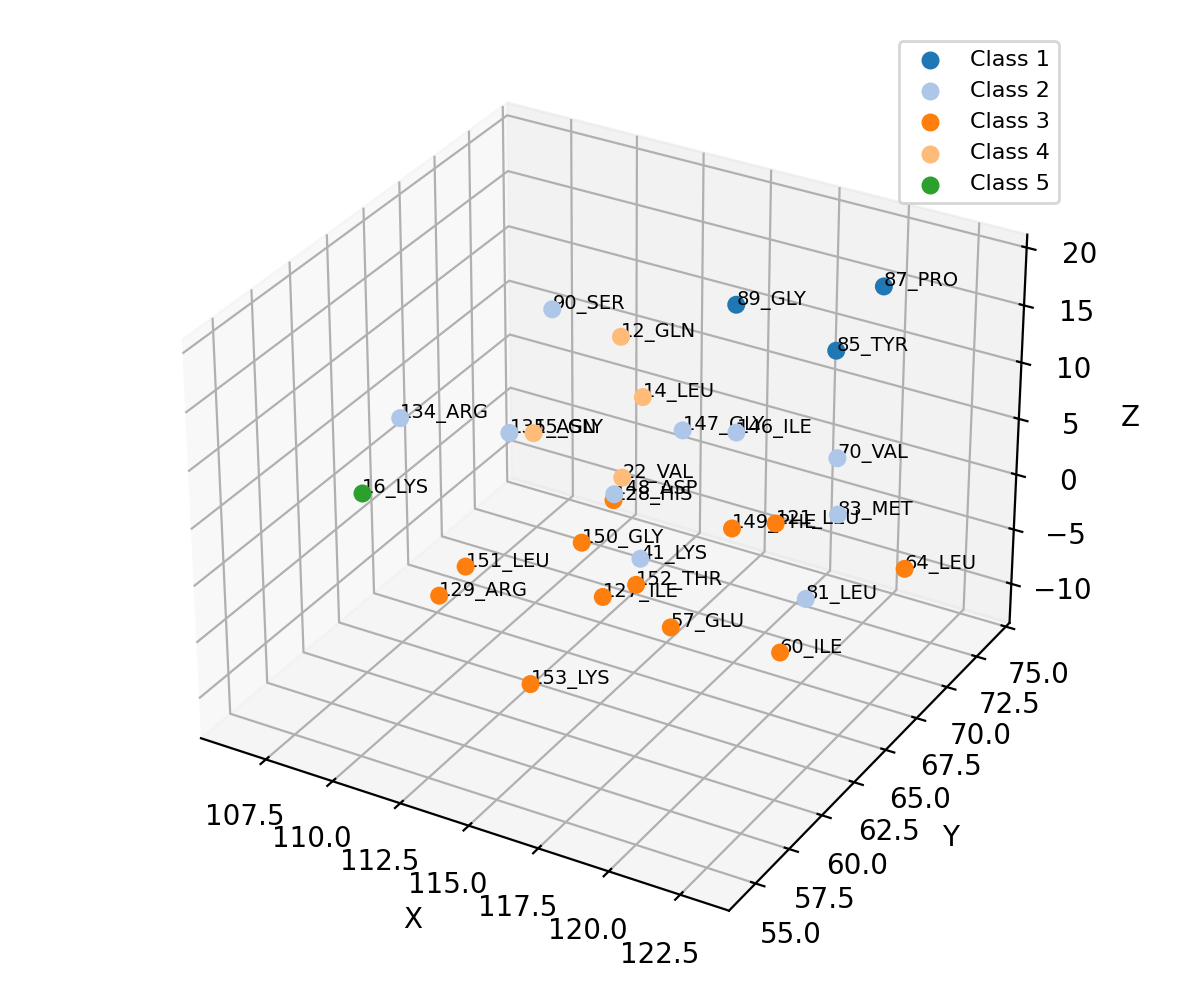}
        \caption{jak2}
    \end{subfigure} &
    \begin{subfigure}[t]{0.18\textwidth}
        \centering
        \includegraphics[width=\linewidth]{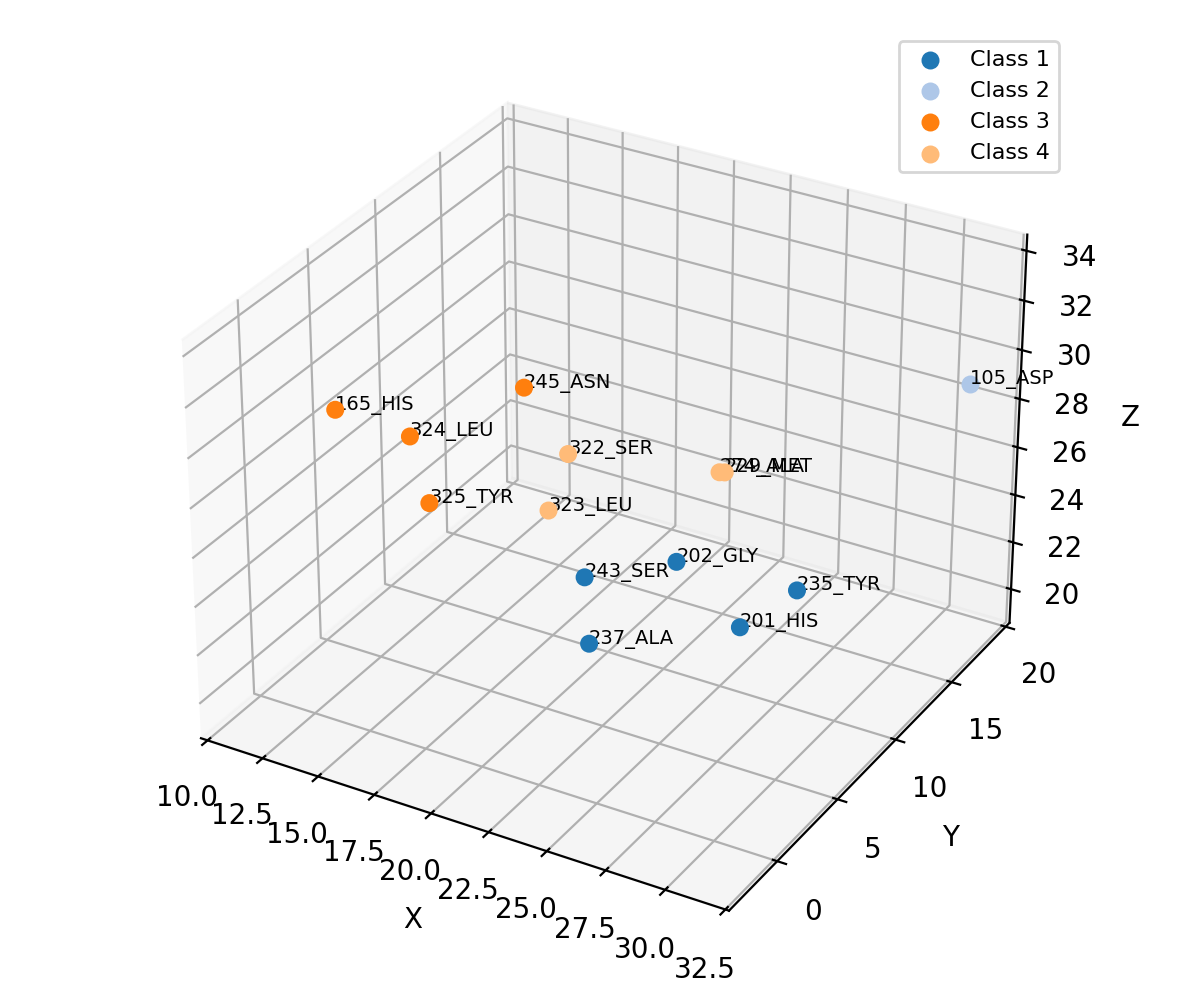}
        \caption{parp1}
    \end{subfigure}

    \end{tabular}

    \caption{Visualization results for five protein targets. The first row shows residue distance matrices, and the second row shows clustered residues (hotspots) for each target.}
    \label{fig:cluster}
\end{figure}

Finally, $k$-means clustering is performed on this embedding space to partition $\mathcal V$ into $m$ disjoint subsets
\(
\{\mathcal S_1, \ldots, \mathcal S_m\},
\)
each representing a spatial hotspot within the binding pocket. These subpockets serve as the structural basis for defining preference-conditioned reward components in Section~\ref{subsec:reward}. The visualization of distance matrices and clustered hotspots are provided in \Cref{fig:cluster}.

\begin{figure*}[t]
    \centering
    \includegraphics[width=\textwidth,height=!]{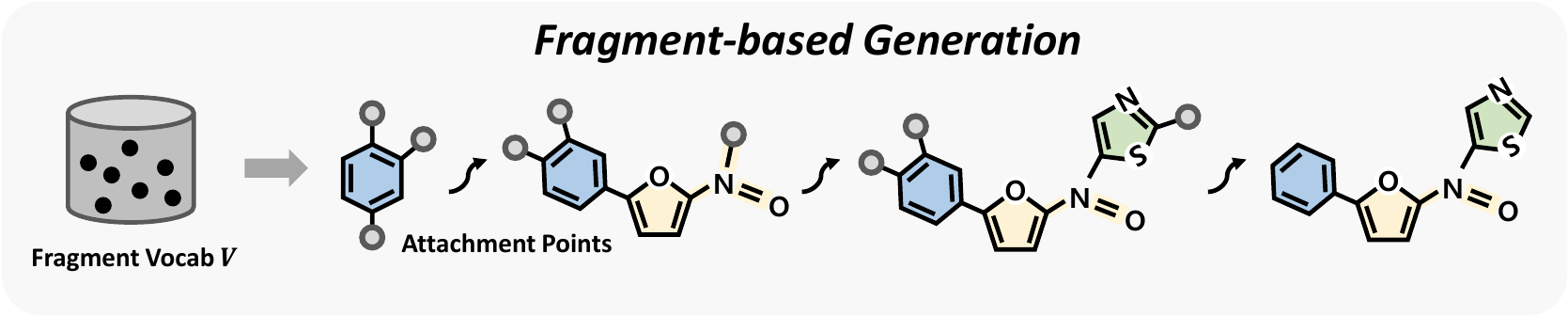}
    \caption{Fragment-based generation sequentially attaches fragments from a fragment vocabulary to a partial graph until termination.}
    \label{fig:generator}
\end{figure*}

\section{Fragment-based Generator}
\label{app:generator}

Following \cite{yang21freed}, we frame the molecular generation as a sequential decision process. The vivid illustration of this fragment-based generator is provided in \Cref{fig:generator}.At time step $t$, the state is the partially built graph $\mathcal{G}_t$. Each transition is achieved by a triplet action
\[
    \mathbf{a}_t = (u_t, f_t, v_t),
\]
where $u_t \in \mathcal{U}(\mathcal{G}_t)$ is an attachment site in the current graph, $f_t \in \mathcal{S}$ is a fragment drawn from the vocabulary $\mathcal{S}$, and $v_t \in \mathcal{V}(f_t)$ is an attachment site on the chosen fragment. Attaching $f_t$ to $\mathcal{G}_t$ at $(u_t,v_t)$ yields $\mathcal{G}_{t+1}$.

Formally, we define three action components governed by sub-policies:
\begin{align}
    \mathbb{P}_1(u_t \mid \mathcal{G}_t) &= \Pi_1\big(\mathcal{H}(\mathcal{G}_t), \mathcal{H}_\text{site}(\mathcal{G}_t)\big), \\
    \mathbb{P}_2(f_t \mid u_t,\mathcal{G}_t) &= \Pi_2\big(\phi(u_t), \Phi(\mathcal{S})\big), \\
    \mathbb{P}_3(v_t \mid f_t,u_t,\mathcal{G}_t) &= \Pi_3\big(\psi(f_t), \psi_\text{site}(f_t)\big),
\end{align}
where $\mathcal{H}(\mathcal{G}_t)\in\mathbb{R}^d$ is a global embedding of the partial graph, $\mathcal{H}_\text{site}(\mathcal{G}_t)=\{h_u : u\in\mathcal{U}(\mathcal{G}_t)\}$ gives embeddings of candidate sites, and $\phi(u_t):=h_{u_t}$ denotes the embedding of the chosen site.  
The fragment vocabulary is represented by $\Phi(\mathcal{S})=\{\phi_b : b\in\mathcal{S}\}$, a collection of precomputed fragment descriptors. Each fragment $f$ is further encoded as a global embedding $\psi(f)\in\mathbb{R}^d$ and its attachment sites as $\psi_\text{site}(f)=\{g_v^{(f)}: v\in\mathcal{V}(f)\}$.  

Here $\Pi_1,\Pi_2,\Pi_3$ are neural scoring functions that map their inputs to logits; applying a softmax yields categorical distributions $\mathbb{P}_1,\mathbb{P}_2,\mathbb{P}_3$ over the respective action spaces. The overall policy can be factorized as
\begin{equation}
    \pi(\mathbf{a}_t \mid \mathcal{G}_t) = 
    \mathbb{P}_1(u_t \mid \mathcal{G}_t)\,
    \mathbb{P}_2(f_t \mid u_t,\mathcal{G}_t)\,
    \mathbb{P}_3(v_t \mid f_t,u_t,\mathcal{G}_t).
\end{equation}

Inspired by \cite{lee2024molecule}, we introduce new fragments by applying graph genetic operations (mutation and crossover) on generated molecules \citep{jensen2019graph} and incorporating novel substructures back into $\mathcal{S}$. When the vocabulary grows too large, only the top-ranked fragments are retained.

\section{Theoretical Proofs and Analysis}
\label{sec:proofs}

This appendix provides a comprehensive theoretical treatment of the energy-based fragment scoring framework introduced in Section~\ref{subsec:ebm}. Our goal is to move beyond heuristic feature importance by grounding our scoring mechanism in the rigorous principles of differential geometry and calculus on manifolds.

Specifically, we address a fundamental question: \emph{How does the discrete removal of a chemical substructure relate to the continuous gradient information learned by the neural network?} To answer this, we model the ablation process as a trajectory through the feature space and analyze the energy dynamics along this path. We rigorously establish that our discrete ablation score is mathematically equivalent to the path integral of gradients (Integrated Gradients), thereby satisfying the axiom of completeness in attribution theory. Furthermore, we provide error bounds to justify the stability of this metric against local perturbations.

\subsection{Preliminaries: Geometry of the Feature Space}

To formalize our analysis, we treat the molecular representation not merely as a matrix of numbers, but as a vector in a high-dimensional Hilbert space.

Let $\mathcal{H} = \mathbb{R}^{|\mathcal{V}| \times d}$ denote the Hilbert space of node features equipped with the Frobenius inner product. For any two feature matrices $\mathbf{A}, \mathbf{B} \in \mathcal{H}$, the inner product is defined as:
\begin{equation}
    \langle \mathbf{A}, \mathbf{B} \rangle_F \coloneqq \text{Tr}(\mathbf{A}^\top \mathbf{B}) = \sum_{i,j} A_{ij} B_{ij}.
\end{equation}
The induced norm $\|\mathbf{A}\|_F = \sqrt{\langle \mathbf{A}, \mathbf{A} \rangle_F}$ provides a measure of the signal magnitude.
The energy function $E_\theta(\mathbf{H}, y)$ maps a point in this feature space (conditioned on property $y$) to a scalar potential. The gradient $\nabla_{\mathbf{H}} E_\theta(\mathbf{H}, y)$ serves as a vector field over $\mathcal{H}$, indicating the direction of steepest energy ascent.

\begin{definition}[Fragment Subspace Projection]
Given a fragment $\mathcal{F} \subseteq \mathcal{V}$, we define the binary indicator mask $\mathbf{1}_{\mathcal{F}} \in \{0, 1\}^{|\mathcal{V}| \times d}$ such that entries corresponding to nodes in $\mathcal{F}$ are 1, and 0 otherwise.
Geometrically, the operation $\mathbf{1}_{\mathcal{F}} \odot \mathbf{H}$ represents the orthogonal projection of the molecule's features onto the subspace spanned by the fragment. Conversely, the ablation is a projection onto the orthogonal complement, realized by $(\mathbf{1} - \mathbf{1}_{\mathcal{F}}) \odot \mathbf{H}$.
\end{definition}

\subsection{Proof of Theorem~\ref{thm:gradient_flow}: Manifold Gradient Equivalence}
\label{app:proof}

One might intuitively expect that the importance of a fragment is proportional to the gradients of its atoms. However, in deep non-linear networks, gradients can saturate or vanish. The "Energy Gap" method we propose (computing $E_{\text{original}} - E_{\text{ablated}}$) captures the global effect of the fragment.
Theorem \ref{thm:gradient_flow} bridges the gap between this global discrete difference and the local continuous gradients.

\textbf{Theorem \ref{thm:gradient_flow}.} \textit{Let $\gamma: [0, 1] \to \mathbb{R}^{|\mathcal{V}| \times d}$ be a linear path parameterizing the feature ablation from $\mathbf{H}$ to $\mathcal{P}_{\mathcal{F}^\perp}(\mathbf{H})$. Assuming $E_\theta$ is continuously differentiable, the discrete energy variation $\mathcal{I}(\mathcal{F})$ is exactly the path integral of the directional derivatives along $\gamma$:}
\begin{equation}
    \mathcal{I}(\mathcal{F}) = - \int_{0}^{1} \left\langle \nabla_{\mathbf{H}} E_\theta(\gamma(t), y), \, \mathbf{1}_{\mathcal{F}} \odot \mathbf{H} \right\rangle_F \, dt.
\end{equation}

\subsubsection{Proof Derivation}

The proof relies on constructing a virtual trajectory connecting the intact molecule to the ablated version and applying the Fundamental Theorem of Calculus along this path.

\begin{lemma}[Path Dynamics]
\label{lemma:path}
We define the ablation path $\gamma(t)$ connecting the original molecule state $\mathbf{H}$ (at $t=0$) to the ablated state (at $t=1$) as:
\begin{equation}
    \gamma(t) = \mathbf{H} - t(\mathbf{1}_{\mathcal{F}} \odot \mathbf{H}), \quad t \in [0, 1].
\end{equation}
The velocity vector (tangent) of this path is constant and points in the negative direction of the fragment features:
\begin{equation}
    \frac{d}{dt} \gamma(t) = - (\mathbf{1}_{\mathcal{F}} \odot \mathbf{H}).
\end{equation}
\end{lemma}

\begin{proof}[Proof of Lemma \ref{lemma:path}]
By definition, the ablated state is $\mathbf{H} - \mathbf{1}_{\mathcal{F}} \odot \mathbf{H}$. The linear interpolation is:
\begin{align}
    \gamma(t) &= (1-t)\mathbf{H} + t(\mathbf{H} - \mathbf{1}_{\mathcal{F}} \odot \mathbf{H}) \\
    &= \mathbf{H} - t(\mathbf{1}_{\mathcal{F}} \odot \mathbf{H}).
\end{align}
Differentiating strictly with respect to the scalar parameter $t$ yields the tangent vector $\gamma'(t) = -(\mathbf{1}_{\mathcal{F}} \odot \mathbf{H})$. This confirms the path is a straight line in $\mathcal{H}$.
\end{proof}

\textit{Proof of Theorem \ref{thm:gradient_flow}:}

Let $g(t) = E_\theta(\gamma(t), y)$ represent the scalar energy evaluated along the ablation trajectory. Since $E_\theta$ is a composition of smooth functions (neural network layers), it is differentiable. We apply the Chain Rule to analyze the rate of energy change.
The derivative of the composite function $g(t)$ is the inner product of the gradient of the outer function (the energy field) and the derivative of the inner function (the path velocity):
\begin{equation}
    \frac{dg}{dt} = \left\langle \nabla_{\gamma(t)} E_\theta(\gamma(t), y), \, \frac{d\gamma}{dt} \right\rangle_F.
\end{equation}
Substituting the result from Lemma~\ref{lemma:path}, the rate of change becomes:
\begin{equation}
    \frac{dg}{dt} = \left\langle \nabla_{\mathbf{H}} E_\theta(\gamma(t), y), \, -(\mathbf{1}_{\mathcal{F}} \odot \mathbf{H}) \right\rangle_F.
\end{equation}
By the linearity of the inner product, we pull the negative sign out:
\begin{equation}
    \frac{dg}{dt} = - \left\langle \nabla_{\mathbf{H}} E_\theta(\gamma(t), y), \, \mathbf{1}_{\mathcal{F}} \odot \mathbf{H} \right\rangle_F.
\end{equation}
The total change in energy, which defines our importance score $\mathcal{I}(\mathcal{F})$, is obtained by accumulating these infinitesimal changes from the initial state to the final ablated state:
\begin{align}
    \mathcal{I}(\mathcal{F}) &= E_\theta(\gamma(1), y) - E_\theta(\gamma(0), y) \\
    &= \int_{0}^{1} \frac{dg}{dt} \, dt \\
    &= - \int_{0}^{1} \left\langle \nabla_{\mathbf{H}} E_\theta(\gamma(t), y), \, \mathbf{1}_{\mathcal{F}} \odot \mathbf{H} \right\rangle_F \, dt.
\end{align}

\begin{remark}[Theoretical Implication]
The integral form derived above corresponds to \emph{Integrated Gradients (IG)} applied to the fragment subspace. Theorem \ref{thm:gradient_flow} proves that our "Energy Gap" score is not a heuristic; it is an exact implementation of IG. This ensures that our method satisfies the \textit{Completeness Axiom}: the sum of attributions equals the total function output difference. Consequently, our method correctly accounts for non-linear saturation effects that would otherwise mislead a simple gradient-based analysis.
\end{remark}

\subsection{Error Analysis and Bounds}

While Theorem~\ref{thm:gradient_flow} provides an exact equality, computing the integral requires evaluating the neural network multiple times along the path, which can be computationally expensive. In practice, one might approximate this using only the gradient at the initial point (the original molecule).
Here, we analyze the theoretical error introduced by such a first-order approximation. This analysis tells us how "trustworthy" the local gradients are for estimating global fragment importance.

\begin{assumption}[Lipschitz Smoothness]
\label{assump:lipschitz}
We assume the energy landscape is smooth. Specifically, the gradient $\nabla_{\mathbf{H}} E_\theta$ is $L$-Lipschitz continuous over the domain $\mathcal{H}$. This means there exists a constant $L > 0$ representing the maximum curvature of the energy surface, such that for any $\mathbf{A}, \mathbf{B} \in \mathcal{H}$:
\begin{equation}
    \|\nabla_{\mathbf{H}} E_\theta(\mathbf{A}, y) - \nabla_{\mathbf{H}} E_\theta(\mathbf{B}, y)\|_F \leq L \|\mathbf{A} - \mathbf{B}\|_F.
\end{equation}
\end{assumption}

\begin{theorem}[Lagrangian Error Bound]
\label{thm:perturbation_bound}
The deviation of the exact discrete importance score from its first-order Taylor approximation is quadratically bounded by the energy norm of the fragment features. Specifically:
\begin{equation}
    \left| \mathcal{I}(\mathcal{F}) - \left( - \langle \nabla_{\mathbf{H}} E_\theta(\mathbf{H}, y), \mathbf{1}_{\mathcal{F}} \odot \mathbf{H} \rangle_F \right) \right| \leq \frac{L}{2} \|\mathbf{1}_{\mathcal{F}} \odot \mathbf{H}\|_F^2.
\end{equation}
\end{theorem}

\begin{proof}
Let $\Delta \mathbf{H} = -(\mathbf{1}_{\mathcal{F}} \odot \mathbf{H})$. By the fundamental theorem of calculus derived in the previous section:
\begin{equation}
    \mathcal{I}(\mathcal{F}) = \int_{0}^{1} \langle \nabla E_\theta(\mathbf{H} + t\Delta \mathbf{H}, y), \Delta \mathbf{H} \rangle_F \, dt.
\end{equation}
We aim to compare this to the linear approximation $\langle \nabla E_\theta(\mathbf{H}, y), \Delta \mathbf{H} \rangle_F$. To do so, we add and subtract the gradient at the origin point inside the integral:
\begin{align}
    \mathcal{I}(\mathcal{F}) &= \int_{0}^{1} \langle \nabla E_\theta(\mathbf{H}, y), \Delta \mathbf{H} \rangle_F \, dt \nonumber \\
    &\quad + \int_{0}^{1} \langle \nabla E_\theta(\mathbf{H} + t\Delta \mathbf{H}, y) - \nabla E_\theta(\mathbf{H}, y), \Delta \mathbf{H} \rangle_F \, dt.
\end{align}
The first term is constant with respect to $t$ and simplifies to the linear approximation:
\begin{equation}
    \int_{0}^{1} \langle \nabla E_\theta(\mathbf{H}, y), -(\mathbf{1}_{\mathcal{F}} \odot \mathbf{H}) \rangle_F \, dt = -\langle \nabla E_\theta(\mathbf{H}, y), \mathbf{1}_{\mathcal{F}} \odot \mathbf{H} \rangle_F.
\end{equation}
Let $\mathcal{R}$ denote the second term (the remainder/error). We bound $|\mathcal{R}|$ using the Cauchy-Schwarz inequality and Assumption \ref{assump:lipschitz}:
\begin{align}
    |\mathcal{R}| &= \left| \int_{0}^{1} \langle \nabla E_\theta(\mathbf{H} + t\Delta \mathbf{H}, y) - \nabla E_\theta(\mathbf{H}, y), \Delta \mathbf{H} \rangle_F \, dt \right| \\
    &\leq \int_{0}^{1} \left\| \nabla E_\theta(\mathbf{H} + t\Delta \mathbf{H}, y) - \nabla E_\theta(\mathbf{H}, y) \right\|_F \| \Delta \mathbf{H} \|_F \, dt \\
    &\leq \int_{0}^{1} L \| t\Delta \mathbf{H} \|_F \| \Delta \mathbf{H} \|_F \, dt \\
    &= L \| \Delta \mathbf{H} \|_F^2 \int_{0}^{1} t \, dt.
\end{align}
Computing the standard scalar integral $\int_{0}^{1} t \, dt = \frac{1}{2}$, and substituting $\Delta \mathbf{H}$, we obtain the final bound:
\begin{equation}
    |\mathcal{R}| \leq \frac{L}{2} \| \mathbf{1}_{\mathcal{F}} \odot \mathbf{H} \|_F^2.
\end{equation}
\end{proof}

\textbf{Interpretation:} The error bound $\frac{L}{2} \|\mathbf{1}_{\mathcal{F}} \odot \mathbf{H}\|_F^2$ implies that the linear approximation is most accurate for (1) "flat" energy landscapes where $L$ is small, or (2) fragments with small feature norms. For large, complex fragments, the non-linearity (captured by $\mathcal{R}$) becomes significant, further justifying our use of the exact Energy Gap method rather than simple gradient maps.

\subsection{Consistency of Gradient-Based Ranking}

Finally, we utilize the previous results to justify the consistency between the rigorous ablation score and simpler gradient-based metrics often used in saliency maps. In many practical scenarios, we are interested in the \textit{ranking} of fragments rather than their exact energy contribution.

\begin{corollary}[Ranking Alignment under Homogeneity]
\label{cor:ranking_alignment}
Suppose the fragment features satisfy a local homogeneity condition where the gradient is aligned with the feature vector. Then, ranking fragments by the magnitude of the energy gap $|\mathcal{I}(\mathcal{F})|$ is asymptotically equivalent to ranking by the summed gradient norms.
\end{corollary}

\begin{proof}
From Theorem~\ref{thm:perturbation_bound}, if the remainder term is small relative to the linear term, we have:
\begin{equation}
    |\mathcal{I}(\mathcal{F})| \approx \left| \langle \nabla_{\mathbf{H}} E_\theta(\mathbf{H}, y), \mathbf{1}_{\mathcal{F}} \odot \mathbf{H} \rangle_F \right|.
\end{equation}
This term can be decomposed into individual node contributions:
\begin{equation}
    \left| \sum_{i \in \mathcal{F}} \nabla_{\mathbf{x}_i} E_\theta(\mathbf{H}, y)^\top \mathbf{x}_i \right|.
\end{equation}
Assuming alignment ($\nabla^\top \mathbf{x} \propto \|\nabla\|$), this is proportional to $\sum_{i \in \mathcal{F}} \|\nabla_{\mathbf{x}_i} E_\theta\|_2$. Thus, the computationally expensive ablation score $\mathcal{I}(\mathcal{F})$ and the efficient gradient norm score induce the same ordering of fragments. This result provides theoretical grounding for using gradient-based screening as a preliminary filter before precise energy-based scoring.
\end{proof}

\section{Limitations and future works}
\label{app:limitations}
While \themodel demonstrates strong performance in exploring diverse binding modes, our current reward formulation relies on simplified interaction counts and static docking poses. First, treating all interaction types equally---without weighting them by energetic contribution---prioritizes interaction diversity over fine-grained affinity modeling. Assigning biochemically accurate weights is non-trivial and target-dependent, yet future iterations could leverage learnable or physics-based weighting schemes to better capture residue-specific importance. Second, like most structure-based generative models, our reliance on rapid static docking (e.g., QVina) may occasionally reward suboptimal or physically strained poses. The PLIP-IFP analyses therefore quantify \emph{docking-predicted} interaction modes rather than experimentally confirmed binding poses, and IFP cluster counts may retain some sample-size dependence despite the complementary entropy statistics. Crystallographic-mode recovery is limited to BRAF and JAK2 because only these targets provide directly comparable paired mode labels. Future work should incorporate receptor ensembles, rigorous energy minimization, and experimental structural validation.

\section{Additional Empirical Results}
\label{app:results}

\subsection{Interaction-level diversity and crystallographic-mode recovery}
\label{app:interaction_eval}
For every qualified docked complex, we construct a binary full-pocket PLIP fingerprint whose features are residue--interaction-type pairs. Pairwise Jaccard distances are clustered with the Butina algorithm at distance cutoff 0.5. We report the number of recurring clusters; entropy $H=-\sum_c p_c\log p_c$, where $p_c$ is the fraction of hits in cluster $c$; and the effective number of modes $\exp(H)$. Cluster count measures breadth, whereas entropy and effective modes also account for population balance. Because cluster count can depend on the number of qualified hits, we interpret all three statistics jointly.

\begin{table}[!htbp]
    \caption{\small \textbf{Full-pocket interaction-fingerprint diversity.} Results are means and standard deviations over three runs. The best result for each target is highlighted in bold.}
    \label{tab:ifp_full}
    \centering
    \resizebox{0.86\textwidth}{!}{
    \begin{tabular}{llccc}
    \toprule
    Target & Method & IFP clusters & IFP entropy $H$ & Effective modes $\exp(H)$\\
    \midrule
    \multirow{5}{*}{PARP1}
      & REINVENT    & $52.33\pm4.04$  & $1.62\pm0.09$ & $5.05\pm0.45$\\
      & FREED       & $102.67\pm6.11$ & $2.47\pm0.08$ & $11.83\pm0.95$\\
      & RationaleRL & $96.33\pm5.51$  & $2.42\pm0.10$ & $11.25\pm1.12$\\
      & GEAM        & $138.67\pm7.02$ & $2.93\pm0.07$ & $18.73\pm1.26$\\
      & \textbf{\themodel} & $\mathbf{192.00\pm6.20}$ & $\mathbf{3.45\pm0.07}$ & $\mathbf{31.40\pm0.96}$\\
    \midrule
    \multirow{5}{*}{FA7}
      & REINVENT    & $21.33\pm3.21$ & $0.77\pm0.09$ & $2.16\pm0.20$\\
      & FREED       & $44.67\pm4.51$ & $1.43\pm0.08$ & $4.18\pm0.34$\\
      & RationaleRL & $41.33\pm4.04$ & $1.38\pm0.10$ & $3.97\pm0.40$\\
      & GEAM        & $61.67\pm5.51$ & $1.85\pm0.07$ & $6.36\pm0.45$\\
      & \textbf{\themodel} & $\mathbf{94.33\pm1.53}$ & $\mathbf{2.37\pm0.06}$ & $\mathbf{10.71\pm0.34}$\\
    \midrule
    \multirow{5}{*}{5HT1B}
      & REINVENT    & $3.67\pm0.58$  & $0.07\pm0.03$ & $1.07\pm0.03$\\
      & FREED       & $6.33\pm0.58$  & $0.17\pm0.04$ & $1.19\pm0.05$\\
      & RationaleRL & $6.00\pm1.00$  & $0.15\pm0.05$ & $1.16\pm0.06$\\
      & GEAM        & $8.33\pm0.58$  & $0.24\pm0.04$ & $1.27\pm0.05$\\
      & \textbf{\themodel} & $\mathbf{11.67\pm0.58}$ & $\mathbf{0.32\pm0.02}$ & $\mathbf{1.37\pm0.03}$\\
    \midrule
    \multirow{5}{*}{BRAF}
      & REINVENT    & $7.33\pm1.53$  & $0.18\pm0.06$ & $1.20\pm0.07$\\
      & FREED       & $13.67\pm2.08$ & $0.41\pm0.08$ & $1.51\pm0.12$\\
      & RationaleRL & $12.67\pm1.53$ & $0.38\pm0.09$ & $1.46\pm0.13$\\
      & GEAM        & $20.33\pm2.52$ & $0.68\pm0.07$ & $1.97\pm0.14$\\
      & \textbf{\themodel} & $\mathbf{35.00\pm1.41}$ & $\mathbf{1.11\pm0.08}$ & $\mathbf{3.03\pm0.08}$\\
    \midrule
    \multirow{5}{*}{JAK2}
      & REINVENT    & $14.33\pm2.08$ & $0.39\pm0.08$ & $1.48\pm0.12$\\
      & FREED       & $31.67\pm3.51$ & $0.97\pm0.09$ & $2.64\pm0.24$\\
      & RationaleRL & $28.67\pm3.06$ & $0.92\pm0.10$ & $2.51\pm0.26$\\
      & GEAM        & $47.33\pm4.04$ & $1.52\pm0.08$ & $4.57\pm0.37$\\
      & \textbf{\themodel} & $\mathbf{77.33\pm2.16}$ & $\mathbf{2.31\pm0.05}$ & $\mathbf{10.10\pm0.35}$\\
    \bottomrule
    \end{tabular}}
\end{table}

For crystallographic-mode recovery, we use BRAF and JAK2 because both provide well-established pairs of type-I/DFG-in and type-II/DFG-out structures within the same target. Specifically, the references are BRAF 4MNF (DFG-in/type-I) and 5HI2 (DFG-out/type-II), and JAK2 6VGL (type-I) and 7TEU (type-II/DFG-out). The remaining targets do not provide equally clear and directly comparable paired labels. For each reference conformation, we re-dock generated hits into that receptor and compare their PLIP IFPs with the native ligand IFP; similarity $\geq0.4$ defines recovery.

\begin{table}[!htbp]
\centering
\caption{\small \textbf{Crystallographic-mode recovery and reference separation.} The left panel reports recovery rates; the right panel compares the two native references for each target using the same full-pocket representation.}
\label{tab:crystal_full}
\begin{minipage}[t]{0.56\textwidth}
\centering\footnotesize\setlength{\tabcolsep}{3pt}
\adjustbox{max width=\linewidth}{\begin{tabular}{llcrrr}
\toprule
Target & Crystal mode & PDB & \themodel & GEAM & FREED\\
\midrule
BRAF & DFG-in/type-I & 4MNF & \textbf{48.9\%} & 20.3\% & 11.7\%\\
BRAF & DFG-out/type-II & 5HI2 & \textbf{12.1\%} & 6.4\% & 0.0\%\\
JAK2 & type-I & 6VGL & \textbf{1.3\%} & 0.0\% & 0.0\%\\
JAK2 & type-II/DFG-out & 7TEU & \textbf{11.5\%} & 5.7\% & 2.8\%\\
\bottomrule
\end{tabular}}
\end{minipage}\hfill
\begin{minipage}[t]{0.40\textwidth}
\centering\footnotesize\setlength{\tabcolsep}{3pt}
\adjustbox{max width=\linewidth}{\begin{tabular}{lcc}
\toprule
 & BRAF & JAK2\\
\midrule
IFP Jaccard & 0.375 & 0.188\\
Shared IFP features & $6/16$ & $3/16$\\
Residue Jaccard & 0.462 & 0.250\\
Shared residues & $6/13$ & $3/12$\\
\bottomrule
\end{tabular}}
\end{minipage}
\end{table}

Both native-pair IFP similarities are below the recovery threshold, so recovery does not merely match a conserved interaction core. Absolute rates remain conservative because training uses a fixed receptor while evaluation uses independent crystal structures with induced-fit, side-chain, DFG-loop, and $\alpha$C-helix differences. These are docking-predicted interaction modes, not experimentally confirmed poses.

\FloatBarrier
\subsection{Matched MORL comparisons and hotspot ablations}
All MORL baselines use the same fragment action space, vector reward, training protocol, docking-oracle budget, and three random seeds as \themodel.

\begin{table}[!htbp]
\centering
\caption{\small \textbf{Matched MORL comparison.} Means and standard deviations over three runs.}
\label{tab:morl_full}
\begin{minipage}[t]{\textwidth}
\centering
\textbf{Novel hit ratio (\%)}\vspace{1mm}

\resizebox{\textwidth}{!}{
\begin{tabular}{lccccc}
\toprule
Method & PARP1 & FA7 & 5HT1B & BRAF & JAK2\\
\midrule
GPI & $37.38\pm1.12$ & $19.84\pm0.91$ & $37.67\pm0.74$ & $27.53\pm1.08$ & $38.31\pm1.16$\\
C-MORL & $38.74\pm0.95$ & $20.92\pm0.83$ & $36.46\pm0.62$ & $27.38\pm0.89$ & $38.27\pm1.04$\\
PreCo & $38.81\pm0.88$ & $21.67\pm0.72$ & $37.93\pm0.51$ & $26.84\pm1.03$ & $39.12\pm0.92$\\
\textbf{\themodel} & $\mathbf{42.62\pm0.73}$ & $\mathbf{24.58\pm0.65}$ & $\mathbf{42.15\pm0.13}$ & $\mathbf{30.77\pm0.97}$ & $\mathbf{44.86\pm0.98}$\\
\bottomrule
\end{tabular}}
\end{minipage}\par\vspace{2mm}
\begin{minipage}[t]{\textwidth}
\centering
\textbf{Novel top-5\% docking score}\vspace{1mm}

\resizebox{\textwidth}{!}{
\begin{tabular}{lccccc}
\toprule
Method & PARP1 & FA7 & 5HT1B & BRAF & JAK2\\
\midrule
GPI & $-10.98\pm0.14$ & $-9.01\pm0.11$ & $-11.02\pm0.16$ & $-10.31\pm0.13$ & $-10.67\pm0.12$\\
C-MORL & $-11.85\pm0.10$ & $-8.92\pm0.08$ & $-11.24\pm0.12$ & $-10.63\pm0.11$ & $-10.51\pm0.09$\\
PreCo & $-11.24\pm0.08$ & $-9.17\pm0.05$ & $-11.61\pm0.11$ & $-10.88\pm0.10$ & $-10.73\pm0.06$\\
\textbf{\themodel} & $\mathbf{-12.93\pm0.06}$ & $\mathbf{-10.19\pm0.02}$ & $\mathbf{-13.03\pm0.10}$ & $\mathbf{-12.50\pm0.09}$ & $\mathbf{-12.01\pm0.02}$\\
\bottomrule
\end{tabular}}
\end{minipage}
\end{table}

\begin{table}[!htbp]
\centering
\caption{\small \textbf{Effect of retaining conserved residues in every hotspot objective.} Novel hit ratio (\%) over three runs.}
\label{tab:full_union}
\resizebox{\linewidth}{!}{
\begin{tabular}{lccccc}
\toprule
Method & PARP1 & FA7 & 5HT1B & BRAF & JAK2\\
\midrule
Full union (including intersection) & $35.23\pm0.63$ & $18.33\pm0.89$ & $35.21\pm0.53$ & $24.21\pm0.96$ & $38.67\pm0.91$\\
\textbf{Non-overlapping residues} & $\mathbf{42.62\pm0.73}$ & $\mathbf{24.58\pm0.65}$ & $\mathbf{42.15\pm0.13}$ & $\mathbf{30.77\pm0.97}$ & $\mathbf{44.86\pm0.98}$\\
\bottomrule
\end{tabular}}
\end{table}

The shared docking component protects affinity-driving core contacts in both variants. Removing the intersection only from the hotspot-specific term reduces correlation among objectives and encourages differentiating subpocket interactions, explaining the consistent advantage in \Cref{tab:full_union}.

\begin{table}[!htbp]
\centering
\caption{\small \textbf{Interaction-level contribution of the hotspot reward on BRAF.}}
\label{tab:hotspot_ablation}
\begin{tabular}{lccc}
\toprule
Method & IFP clusters & IFP entropy $H$ & Effective modes\\
\midrule
\themodel-h & $25.67\pm1.53$ & $0.89\pm0.03$ & $2.20\pm0.08$\\
\textbf{\themodel} & $\mathbf{35.00\pm1.41}$ & $\mathbf{1.11\pm0.08}$ & $\mathbf{3.03\pm0.08}$\\
\bottomrule
\end{tabular}
\end{table}

\FloatBarrier
\subsection{Sensitivity to energy-model dataset size}
\label{app:data_sensitivity}
We train the energy model on stratified 10\%, 25\%, 50\%, 75\%, and 100\% subsets of the empirical BRAF and JAK2 distributions, preserving the property-label distribution in each subset. We then evaluate the downstream fragment vocabulary under the same generation protocol.

\begin{table}[!htbp]
\centering
\caption{\small \textbf{Dataset-size sensitivity of energy-based fragment scoring.} Means and standard deviations over three runs.}
\label{tab:data_sensitivity}
\resizebox{\textwidth}{!}{
\begin{tabular}{llccccc}
\toprule
Metric & Target & 10\% & 25\% & 50\% & 75\% & 100\%\\
\midrule
\multirow{2}{*}{Novel hit ratio (\%)}
 & BRAF & $15.24\pm0.76$ & $20.83\pm0.84$ & $26.17\pm0.91$ & $29.24\pm0.95$ & $\mathbf{30.77\pm0.97}$\\
 & JAK2 & $30.76\pm1.03$ & $34.68\pm1.01$ & $40.02\pm0.99$ & $43.46\pm0.98$ & $\mathbf{44.86\pm0.98}$\\
\midrule
\multirow{2}{*}{Novel top-5\% docking score}
 & BRAF & $-10.35\pm0.04$ & $-11.14\pm0.06$ & $-11.89\pm0.07$ & $-12.28\pm0.08$ & $\mathbf{-12.50\pm0.09}$\\
 & JAK2 & $-10.08\pm0.04$ & $-10.76\pm0.03$ & $-11.43\pm0.03$ & $-11.81\pm0.02$ & $\mathbf{-12.01\pm0.02}$\\
\bottomrule
\end{tabular}}
\end{table}

Performance improves and begins to saturate with more target-relevant data. This supports the intended interpretation that such data are beneficial rather than an all-or-nothing prerequisite: for underexplored targets, a virtually screened public library can supply the scored empirical distribution before directed generation.

\begin{table*}[h!]
    \caption{\small \textbf{GuacaMol MPO \#Circles results.} The results are the means of 3 runs. The best and second-best results are highlighted in bold and underlined, respectively.}
    \centering
    \resizebox{\textwidth}{!}{
    \renewcommand{\arraystretch}{1}
    \renewcommand{\tabcolsep}{1.5mm}
    \begin{tabular}{lcccccccc}
    \toprule
        \multirow{2.5}{*}{Method} & \multicolumn{7}{c}{Benchmark} \\
    \cmidrule(l{2pt}r{2pt}){2-8}
        & Amlodipine & Fexofenadine & Osimertinib & Perindopril & Ranolazine & Sitagliptin & Zaleplon \\
    \midrule
        REINVENT~\citep{olivecrona2017molecular} & 303.7 & 343.3 & \underline{452.3} & 318.3 & 253.3 & \textbf{398.3} & 275.3 \\
        Graph GA~\citep{jensen2019graph} & 258.7 & 333.3 & 270.3 & 278.7 & \underline{364.7} & 306.3 & 272.7 \\ 
        STONED~\citep{nigam2021beyond} & 303.7 & 330.3 & 301.3 & 301.0 & 316.7 & 326.3 & 280.3 \\
        GEAM-static \citep{lee2024molecule} & 412.0 & 397.7 & 315.3 & 318.0 & 256.7 & 233.0 & 267.0 \\
        GEAM \citep{lee2024molecule} & \underline{424.0} & \underline{502.0} & 435.0 & \textbf{377.3} & 295.3 & 257.0 & \underline{336.0} \\
        \midrule
        \themodel (Ours) & \textbf{480.5} & \textbf{542.1} & \textbf{475.5} & \underline{374.8} & \textbf{405.2} & \underline{352.7} & \textbf{362.6} \\
    \bottomrule
    \end{tabular}}
    \label{tab:mpo_novel}
\end{table*}

\FloatBarrier
\subsection{Sensitivity of preference sampling size} 
As mentioned in \Cref{subsec:env_sac}, the maximization over $\bm\omega_i$ requires approximation. It is sampled by first drawing i.i.d. standard normal vectors from $\bf\mathcal{N}(0, I)$, applying an element-wise absolute value to ensure non-negativity, and then performing $\ell_1$ normalization. This ensures that the resulting vectors are uniformly distributed on the probability simplex. These weights are then replicated across the batch dimension to enable efficient preference-conditioned training. 
However, since replicating preference vectors for a batch of PyG graphs increases memory usage, very dense sampling would significantly increase the GPU footprint. In practice, we therefore sample $\bm\omega_i$ five times, which provides a good balance between computational efficiency and performance. To further demonstrate the effect of sampling size $k$, we have conducted a sensitivity analysis and the results are shown in \Cref{tab:sample_size}.

\begin{table}[h]
    \centering
    \caption{Sensitivity analysis of preference sampling size $k$.}
    \begin{adjustbox}{max width=0.7\textwidth} 
     \setlength{\tabcolsep}{12pt}
    \begin{tabular}{lccccc}
        \toprule[1.2pt]
        Novel hit ratio ($\%$)      & parp1       & fa7   & 5ht1b  & braf  & jak2      \\ 
        \midrule
        $k=1$  & 35.025 & 18.277 & 36.272 & 19.333 & 39.047 \\
$k=3$  & 38.903 & 21.667 & 39.622 & 25.887 & 42.887 \\
$k=5$ (\themodel)  & 42.617 & 24.580 & 42.149 & 30.767 & 44.964 \\
$k=10$ & 43.622 & 24.891 & 42.277 & 31.047 & 45.219 \\
$k=20$ & 45.667 & 25.891 & 44.239 & 33.127 & 46.975 \\
        \bottomrule[1.2pt]
        \label{tab:sample_size}
    \end{tabular}
    \end{adjustbox}
\end{table}
It can be observed that with samping size increase, the performance of novel hit ratio on five targets steadily increase. However, due to the memory burden, we set sampling size to $k=5$ as the final configuration to achieve the trade-off between performance and efficiency.

\subsection{Molecular weight distribution analysis}
\begin{figure}[h]\color{Maroon}
    \centering    \includegraphics[width=\textwidth,height=!]{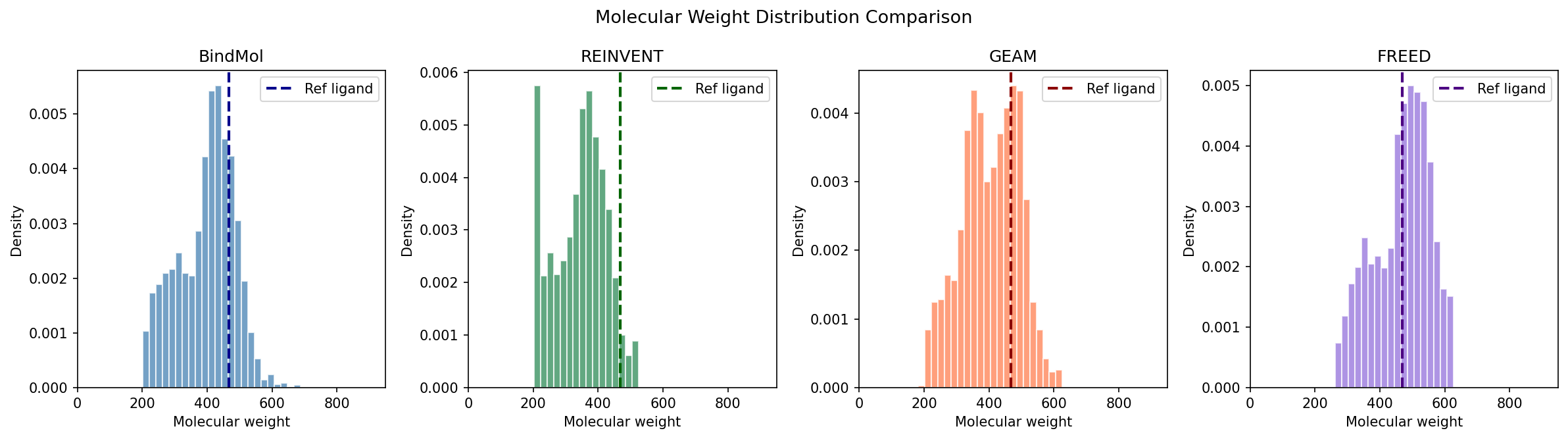}
    \caption{The comparison of molecular weight of generated molecules on jak2 target using different competitive generative approaches.}
    \label{fig:mw}
    \vspace{-3mm}
\end{figure}

To investigate whether our per-residue interaction rewards unintentionally incentivize the generation of larger molecules, we conducted a comparative analysis of the molecular weight (MW) distributions produced by \themodel and several representative baselines. For each method, we sample an equal number of generated molecules for the target jak2 and compute their MW values using RDkit. As an external reference, we also included the MW of experimentally validated ligand extracted from the PDB (PDB id: 7F7W). The results are shown in \Cref{fig:mw}.

As shown in \Cref{fig:mw}, the MW distribution of \themodel closely aligns with those of the baseline methods and is broadly consistent with the distribution observed in PDB ligands. These results indicate that our per-hotspot interaction rewards do not systematically drive the model toward generating heavier molecules. The similarity between generated molecules and reference ligands further suggests that we operate within a chemically reasonable MW range that is typical for bioactive small molecules.

\begin{figure}[h!]
    \centering
    \includegraphics[width=0.9\linewidth]{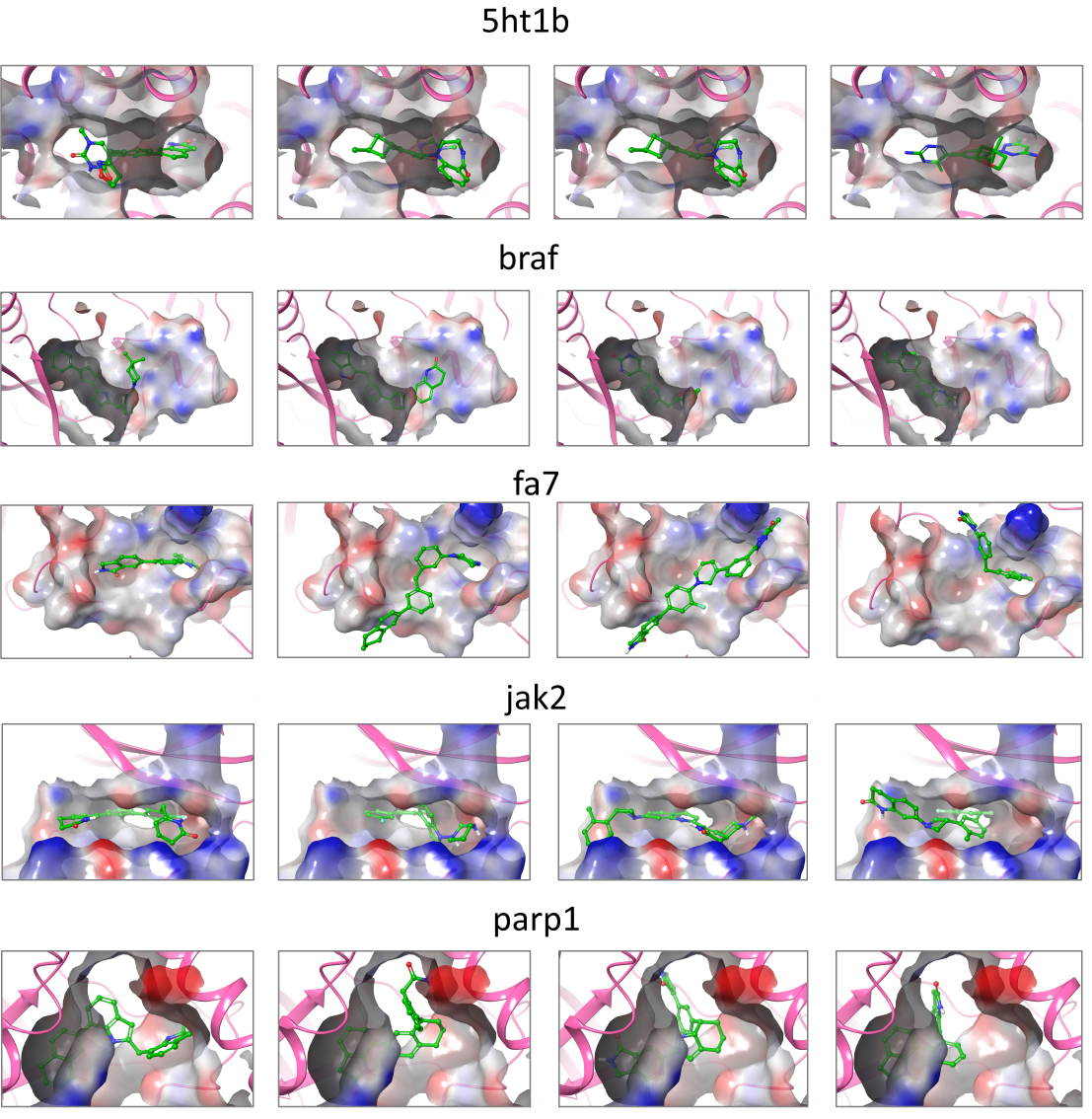}
    \caption{3D visualization of generated binders of five targets}
    \label{fig:pose}
\end{figure}
\clearpage

\clearpage
\label{app:vis}
\begin{figure}[h!]
\vspace{-2mm}
    \centering
    \includegraphics[width=\linewidth]{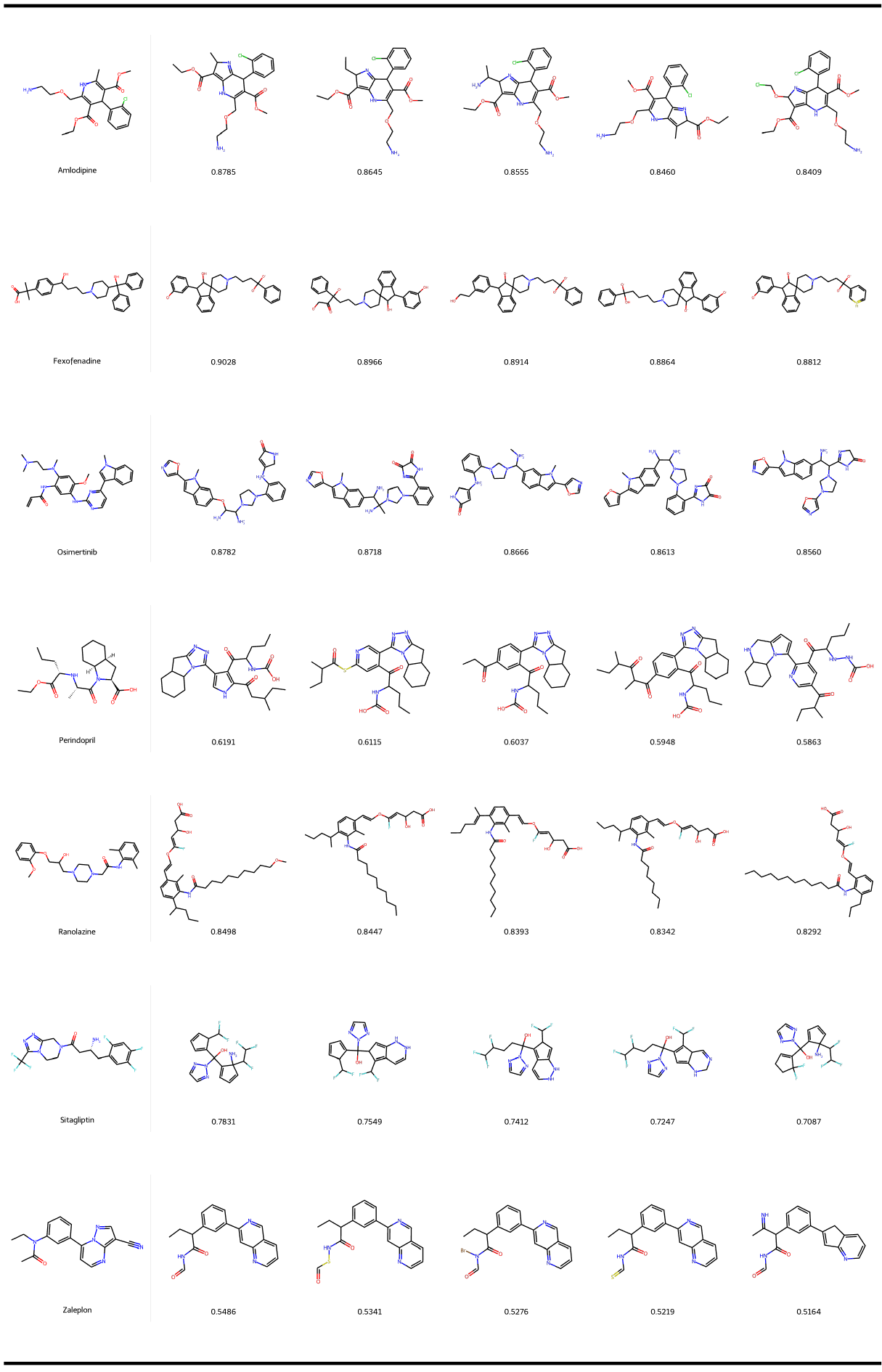}
    \caption{\small The rediscovery molecules of the GuacaMol MPO tasks and the examples of the generated top-5 molecules. The scores are provided at the bottom of each generated molecule.}
    \label{fig:mpo}
\end{figure}
\clearpage


\end{document}